\documentclass[sn-basic]{sn-jnl}

\usepackage{tikz, graphicx}%
\usepackage{multirow}%
\usepackage{amsmath,amssymb,amsfonts}%
\usepackage{amsthm}%
\usepackage{mathrsfs}%
\usepackage[title]{appendix}%
\usepackage{xcolor}%
\usepackage{textcomp}%
\usepackage{manyfoot}%
\usepackage{booktabs}%
\usepackage{algorithm}%
\usepackage[noend]{algpseudocode}%
\usepackage{listings}%
\usepackage{array}
\usepackage{cleveref}
\usepackage{hyperref}
\DeclareMathOperator*{\argmax}{\mathrm{arg\,max}}
\DeclareMathOperator*{\argmin}{\mathrm{arg\,min}}

\theoremstyle{thmstyleone}%
\theoremstyle{thmstyletwo}%
\theoremstyle{thmstylethree}%
\newcommand{\defemph}[1]{\textbf{#1}}
\Crefname{equation}{Eq.}{Eqs.}
\Crefname{section}{Sec.}{Secs.}
\providecommand{\inner}[2]{#1\trans#2} 
\providecommand{\with}{\!\mid\,}
\providecommand{\zeronorm}[1]{\|#1\|_0}
\providecommand{\onenorm}[1]{\|#1\|_1}

\providecommand{\trans}{^\mathrm{T}}
\providecommand{\step}{\mathbf{1}}
\providecommand{\bq}{\mathbf{q}}
\providecommand{\bw}{\mathbf{w}}
\providecommand{\bW}{\mathbf{W}}

\providecommand{\bx}{\mathbf{x}}
\providecommand{\bt}{\mathbf{t}}
\providecommand{\bg}{\mathbf{g}}
\providecommand{\bQ}{\mathbf{Q}}
\providecommand{\by}{\mathbf{y}}
\providecommand{\bX}{\mathbf{X}}
\providecommand{\bbeta}{\boldsymbol{\beta}}

\providecommand{\bbR}{\mathbb{R}}
\providecommand{\setQ}{\mathcal{Q}}

\providecommand{\setI}{\mathcal{I}}
\providecommand{\loss}{\ell}
\providecommand{\Risk}{\mathcal{R}}
\providecommand{\obj}{\mathrm{O}}

\providecommand{\log}{\mathrm{log}}
\providecommand{\sqr}{\mathrm{sqr}}

\providecommand{\halfspace}{\hspace{0.1em}}

\providecommand{\extratiny}[1]{\scalebox{0.8}{#1}}
\providecommand{\extratinytiny}[1]{\scalebox{0.72}{#1}}
\providecommand{\given}{\,|\,}
\providecommand{\from}{\!:\,}

\providecommand{\signg}{\zeta}
\providecommand{\inft}{\ensuremath{\infty}}
\providecommand{\lomedup}[3]{\extratinytiny{#1} #2 \extratinytiny{#3}}
\providecommand{\rev}[1]{{#1}}

\providecommand{\seplen}{\texttt{\color{orange}sep\_length}}
\providecommand{\sepwid}{\texttt{\color{green}sep\_width}}
\providecommand{\petlen}{\texttt{\color{red}pet\_length}}
\providecommand{\petwid}{\texttt{\color{blue}pet\_width}}
\newcommand{\expect}{\mathbb{E}}
\newcommand{\bPhi}{\mathbf{\Phi}}
\newcommand{\bH}{\mathbf{H}}
\newcommand{\bLambda}{\boldsymbol{\Lambda}}
\providecommand{\medinc}{\texttt{\color{orange}med\_income}}
\providecommand{\aveoccup}{\texttt{\color{teal}ave\_occup}}
\providecommand{\lat}{\texttt{\color{red}latitude}}
\providecommand{\lon}{\texttt{\color{blue}longitude}}
\providecommand{\houseage}{\texttt{\color{purple}hs\_age}}
\providecommand{\avebed}{\texttt{\color{cyan}ave\_bedrms}}
\providecommand{\averoom}{\texttt{\color{magenta}ave\_rooms}}
\providecommand{\ReLU}{\textbf{ReLU}}
\begin{document}
\title[Sparse Oblique Rule Boosting for Additive Rule Ensembles]{Sparse Oblique Rule Boosting for Simpler\\ Additive Rule Ensembles}


\author*[1]{\fnm{Shahrzad} \sur{Behzadimanesh}}\email{shahrzad.behzadimanesh@monash.edu}

\author[1]{\fnm{Pierre} \sur{Le Bodic}}\email{pierre.lebodic@monash.edu}

\author[1]{\fnm{Geoffrey} \sur{I. Webb}}\email{geoff.webb@monash.edu}

\author*[2,1]{\fnm{Mario} \sur{Boley}}\email{mboley@is.haifa.ac.il}
\affil*[1]{\orgdiv{Department of Data Science and Artificial Intelligence}, \orgname{Monash University}, \orgaddress{\city{Melbourne}, \state{VIC}, \country{Australia}}}

\affil[2]{\orgdiv{Department of Information Systems}, \orgname{University of Haifa}, \orgaddress{\city{Haifa}, \country{Israel}}}

\abstract{Small additive ensembles of symbolic rules offer interpretable prediction models. Traditionally, these ensembles use rule conditions based on conjunctions of simple threshold propositions $x \geq t$ on a single input variable $x$ and threshold $t$, resulting geometrically in axis-parallel polytopes as decision regions. While this form ensures a high degree of interpretability for individual rules and can be learned efficiently using the gradient boosting approach, it relies on having access to a curated set of expressive input features so that a small ensemble of axis-parallel regions can describe the target variable well. Absent such features, reaching sufficient accuracy requires increasing the number and complexity of individual rules, which diminishes the interpretability of the model.
Here, we extend classical rule ensembles by introducing logical propositions with learnable sparse linear transformations of input variables, i.e., propositions of the form $\bx\trans\bw \geq t$, where $\bw$ is a learnable sparse weight vector, enabling decision regions as general polyhedrons with oblique faces. We propose a learning method using gradient boosting based on a weighted logistic regression. 
Empirical results across 14 regression and classification tasks demonstrate that the proposed method achieves lower model complexity than competitive baselines while maintaining similar or better predictive accuracy. Hence, the approach provides a favorable trade-off between  interpretability and accuracy and reduces the reliance on manual feature engineering.}

\maketitle
\section{Introduction}\label{sec_intro}
\defemph{Additive rule ensembles} are an interpretable and often accurate type of prediction model that corresponds to a set of if-then rules \citep[see, e.g.,][for an introduction]{furnkranz1999separate}.
More technically, one can consider them as \defemph{probabilistic additive models}~\citep[see][]{lafferty1999additive, friedman2003importance} that describe the conditional distribution of an output variable $Y \given X=\bx$ given an input $\bx \in \bbR^d$ through an additive \defemph{model function}
\begin{equation}
    \label{eq:fx}
    f(\bx) = \beta_0 + \beta_1 \varphi_1(\bx) \rev{+} \dots \rev{+} \beta_r \varphi_r(\bx) \enspace ,
\end{equation}
with real-valued coefficients $\beta_0,\dots,\beta_r$ and \textbf{feature functions} $\varphi_j \from \bbR^d \to \bbR$ that map the joint vector of input variable values to real numbers.
In this view of additive rule ensembles, established in~\citet{friedman2008predictive} and \cite{dembczynski2010ender}, the coefficients correspond to the rule consequents---with $\beta_0$ being the consequent of a ``default rule'' that is satisfied for all inputs---and the feature functions correspond to the rule conditions $q \from \bbR^d \to \{0, 1\}$ that have a Boolean output. 
Those conditions are typically given as products or conjunctions of \defemph{univariate threshold conditions}
\begin{align}
q(\bx) &= p(\bx; v_1, t_1, s_1) \dots p(\bx; v_l, t_l, s_l)\label{eq:q}\\
p(\bx; v, t, s) &= \step(sx_v \geq t)\label{eq:p_axis} \enspace .
\end{align}
where individual conditions can be described by a variable index $v$, a threshold $t$, and a sign $s$ that determines the direction of the encoded inequality.
\begin{figure}[t!]
  \centering
  \begin{tikzpicture}
    \node[anchor=south west, inner sep=0] (A) at (1cm,0)
      {\includegraphics[width=0.4\textwidth, trim=1.5cm 0.8cm 0cm 1.7cm, clip]{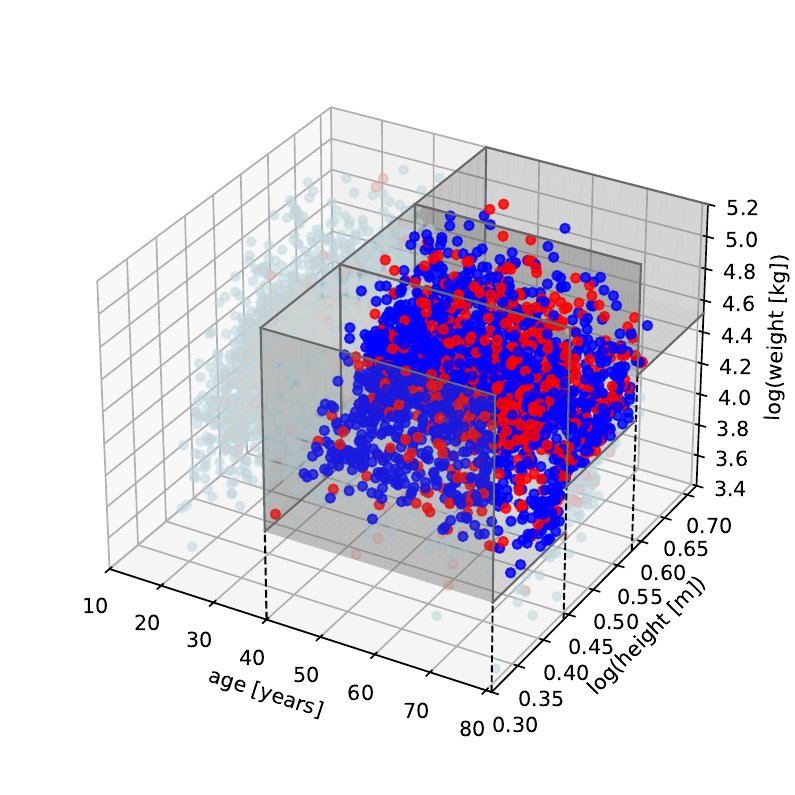}};
    \node[anchor=south east, inner sep=0] (B) at (\textwidth,0)
      {\includegraphics[width=0.44\textwidth, trim=1.5cm 0.8cm 0cm 1.7cm, clip]{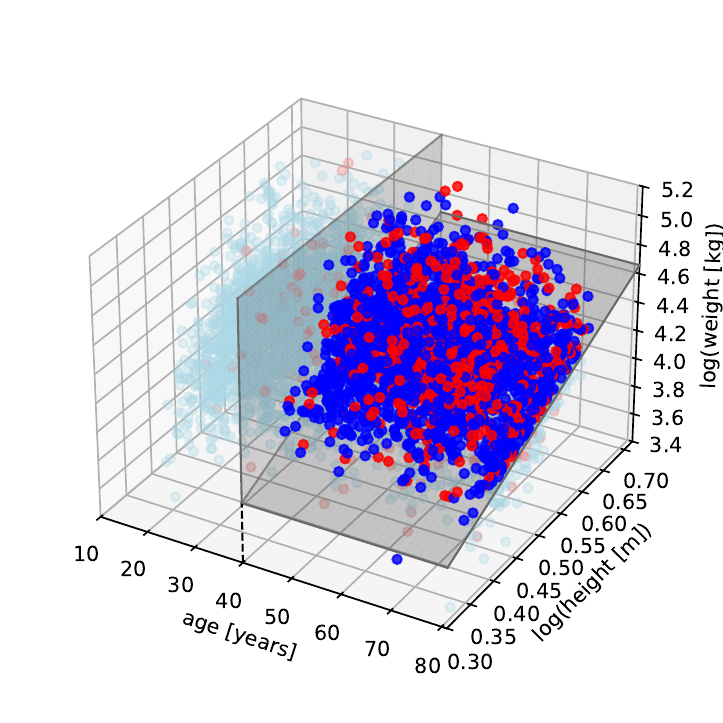}};

    \node[anchor=north west, xshift=-1cm, yshift=1.5cm] at (A.north west)
      {
      \begin{footnotesize}
      $\begin{aligned}
     -3.55 &\textbf{ if }\\
     +2.02 &\textbf{ if } \texttt{age} \geq 40 \textbf{ \& } \texttt{weight} \geq 52.4 \textbf{ \& } 1.6 \geq \texttt{height} \geq 1.4\\
     +2.26 &\textbf{ if } \texttt{age} \geq 40 \textbf{ \& } \texttt{weight} \geq 69.3 \textbf{ \& } 1.8 \geq \texttt{height} \geq 1.6\\
     +2.19 &\textbf{ if } \texttt{age} \geq 40 \textbf{ \& } \texttt{weight} \geq 91.7 \textbf{ \& } 2.1 \geq \texttt{height} \geq 1.8\\
   \end{aligned}$
   \end{footnotesize}
   };
    \node[anchor=north west, xshift=5mm, yshift=8.5mm] at (B.north west)
      {
      \begin{footnotesize}
      $\begin{aligned}
       -2.97 &\textbf{ if }\\
       +1.82 &\textbf{ if } \rev{\texttt{age} \geq 40 \textbf{ \& }}\texttt{weight}/\texttt{height}^2 \geq 25
      \end{aligned}$
      \end{footnotesize}
      };
  \end{tikzpicture}
  \caption{\begin{small}Standard axis-parallel rule ensemble for \rev{modeling} propensity to diabetes based on the 2017-18 CDC NHANES survey \citep{nhanes2020} (\textbf{left}) versus \rev{the proposed} simple polyhedron conditions (\textbf{right}). \rev{In conjunction with} log-transforms the proposed \rev{model} enables learning \rev{ratio conditions, which are used} here \rev{to represent} the concept of body-mass-index---leading to approx. identical log loss with a simpler rule ensemble.\end{small}}
  \label{fig:bmi_example}
  \vspace{-0.3cm}
\end{figure}
Such conditions geometrically correspond to axis-parallel polyhedrons in the input space, and it is this form that endows rule ensembles with their potential interpretability.

Threshold conditions are easily computable for human interpreters in their mind, and the overall ensemble output results from a transparent aggregation over this operation: products/conjunction and sums/disjunctions---properties that are referred to as \textit{simulatability} and \textit{modularity} in the interpretable and explainable AI literature~\citep[see, e.g.,][]{murdoch2019definitions, vollert2021interpretable}.
That being said, the actual interpretability of a rule ensemble is in no small degree determined by its length or complexity, referring to both the number and the length of individual rules.
Indeed, even tree ensembles like random forests or gradient boosted trees can be expressed through \eqref{eq:fx}--\eqref{eq:p_axis}, yet they are typically not considered interpretable due to their large scale.
This is why there are various ways to reduce the complexity of such models while approximately retaining their accuracy.
One of them can be described as \textit{generate-and-select}, where one employs post-processing to select a small subset of initially generated rules, whether through a random forest~\citep{friedman2008predictive, benard2021interpretable} or some other means \citep[e.g.,][]{lakkaraju2016interpretable,wang2017bayesian}.
However, this approach typically suffers from the initial collection having sub-optimal building blocks, and does usually yield inferior results when compared to fitting rules directly one at a time~\citep{dembczynski2010ender, boley2021better, yang2024orthogonal}. This is the approach taken by \textit{rule boosting}, i.e., the application of gradient boosting~\citep{mason1999functional, friedman2001greedy} to individual rule learners as base learners.

In this work, we are interested in another, complementary, approach to improve the accuracy/simplicity trade-off that works on the level of the model definition:
namely to extend the expressiveness of rule conditions from axis-parallel as in \Cref{eq:q,eq:p_axis} to \emph{polyhedral conditions} based on \defemph{oblique hyperplane propositions}, i.e.,
\begin{align}
    f(\bx) &= \beta_0 + \beta_1 q(\bx; \bW_1, \bt_1) + \dots + \beta_r q(\bx; \bW_r, \bt_r) \label{eq:fx_poly}\\ 
    q(\bx; \bW, \bt) &= p(\bx; \bw_1, t_1)\cdots p(\bx; \bw_l, t_l)\label{eq:q_poly}\\
    p(\bx; \bw, t) &= \step(\bw\trans\bx \geq t) \label{eq:p_poly}\enspace ,
\end{align}
where in~\eqref{eq:q_poly} the $\bw_1\dots, \bw_l$ refer to the row vectors of parameter matrix $\bW \in \bbR^{l \times d}$. This is a generalization of the axis-parallel model, which can be recovered by setting the weight matrix entries $w_{i,v_i}$ to $s_i$ (and $0$ elsewhere).

\begin{figure}[t!]
    \centering
    \includegraphics[width=1\textwidth]{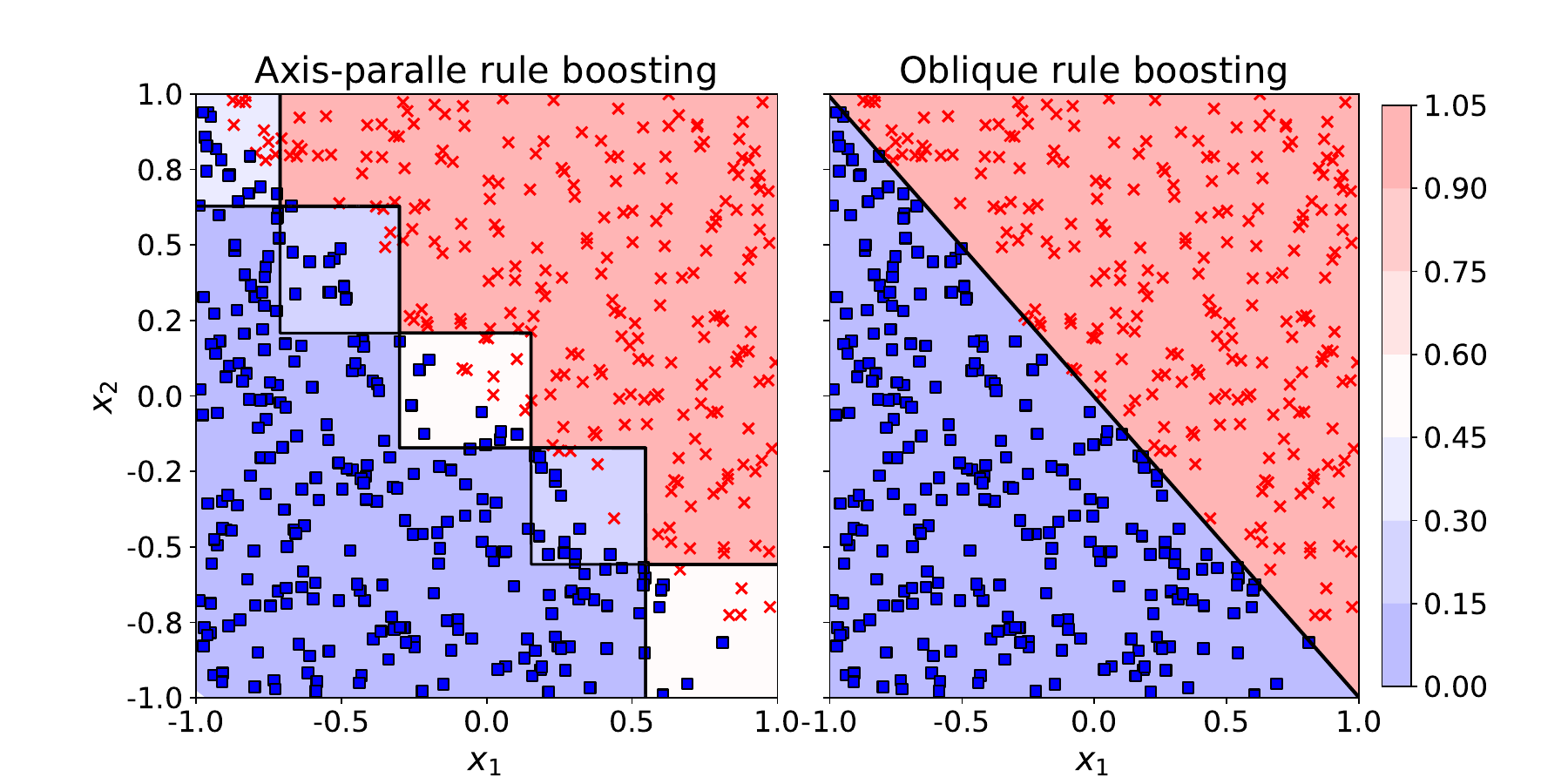}
     \vspace{-0.3cm}
    \caption{Decision regions of rules generated by linear transformation learning and traditional gradient boosting for synthetic data with ground truth decision boundary given by $x_1+x_2=0$.}
    \label{fig:rules_plots}
\end{figure}

\rev{Intuitively,} as shown in \Cref{fig:bmi_example,fig:rules_plots}, this approach can substantially reduce the model complexity required to achieve a certain accuracy, even if one takes into account the increased complexity of individual conditions.
\rev{To make this intuition more precise, we measure} the \defemph{rule ensemble complexity} as
\begin{equation}\label{eq:c}
    C(f) = r + l_1 + \dots + l_r + \|\bW_1\|_0 + \dots + \|\bW_r\|_0\enspace 
\end{equation}
where $l_i$ is the number of rows of $\bW_i$, i.e., the number of propositions in the $i$-th rule, \rev{and $\|\bW_i\|_0$ is the number of non-zero-entries in $\bW_i$, i.e., the total number of variables that contribute to the inequality conditions in rule $i$}. \rev{This measure generalizes similar measures used for axis-parallel rules \citep[e.g.,][]{lakkaraju2016interpretable, yang2024orthogonal}. In the context of oblique rules, it is important to point out that the measure primarily captures the \emph{memory cost} of storing the model rather the cognitive effort of simulating it.
In particular, it does not reflect the increased difficulty of mental arithmetic relative to simple threshold comparisons.
This aspect could be more appropriately captured by weighting the terms in \Cref{eq:c}, and this could be incorporated into the method presented here. However, since determining appropriate weights is beyond the scope of this work, we use the unweighted form for the sake of simplicity.}

\rev{Allowing oblique conditions equips} rule ensembles with a linear, thus \rev{principally} interpretable, form of \rev{representation} learning, which, through log transforms, also enables the automatic learning of product and ratio interaction features (see \Cref{fig:bmi_example}). 
\rev{It is} the same approach taken by oblique \rev{trees}~\citep[see, e.g.,][]{murthy1994system,irsoy2012soft,xu2022one} or ensemble forest models based on those.
However, using oblique trees or forest directly again suffers from a too large complexity and combining them with a post-hoc selection step is likely to suffer from the same inefficiency when starting from axis-parallel trees.
Furthermore, \rev{current} oblique tree \rev{methods} do not offer a \rev{computationally} efficient way to create \emph{sparse} linear transformations.

Therefore, \rev{this work presents and evaluates a method to learn} sparse polyhedral conditions directly within the rule boosting framework without any substantial computational overhead.
In particular, we show how individual oblique cut optimization in gradient boosting reduces to a weighted sparse logistic regression problem where the labels correspond to gradient signs and weights correspond to gradient magnitudes.
Based on this building block, we \rev{develop a search strategy} to find the optimal polyhedral rule condition in a given boosting round.
Finally, we show empirically that the \rev{overall approach} leads to a significantly improved accuracy/simplicity trade-off compared to not only axis-parallel rule boosting but also to an adapted generate-and-filter approach based on forests of oblique trees as well as a recently proposed neuro-symbolic rule learning approach~\citep{dawer2020neural}(see \Cref{table:iris_rules}).

\begin{table}[t]
\centering
\caption{Rules generated by different methods on the Iris dataset (label ``Versicolor'' vs rest). Each ensemble corresponds to the smallest that achieves at least a 50\% reduction in normalized test risk.}
\label{table:iris_rules}
    \begin{tabular}{p{0.05\textwidth}p{0.02\textwidth}p{0.82\textwidth}}
    \toprule
    \multicolumn{3}{l}{\textbf{Sparse Logistic Linear Transformation Boosting}} \\
        -4.54 & if & True \\
        +6.84 & if & (0.86\halfspace \petlen -\sepwid $\geq$ -0.5) AND (\petwid $\leq $1.8) AND (\petlen $\leq$ 5.2)  \\
    \addlinespace[1.5ex]
    \multicolumn{3}{l}{\textbf{Rule Fit Oblique Trees}}\\
        -2.69 & if & True \\  
        -1.8 & if & (-0.9\halfspace \seplen+2.29\halfspace\sepwid -0.06\halfspace\petlen +1.16\halfspace \petwid $>$ 2.6)\\

        +2.39 & if & (0.78\halfspace\sepwid-0.56\halfspace\petlen +1.31\halfspace\petwid $>$ 1.7) AND \\
        & &(0.32\halfspace\sepwid-0.57\halfspace\petlen+1.18\halfspace\petwid $>$ 0.19)\\

        +7.13 & if & (-0.01\halfspace\seplen-1.15\halfspace\sepwid+1.31\halfspace\petwid $>$ -2.1) AND\\
        && (\petlen $\leq$ 5.06) AND (\petwid $\leq$ 1.75) \\
        -2.28 & if & (-0.57\halfspace\petlen+1.16\halfspace\petwid $>$ -0.8)\\
    \addlinespace[1.5ex]
    \multicolumn{3}{l}{\textbf{Traditional Gradient Boosting}} \\
        +0.06 & if & True \\
        -2.84  & if & (\petwid $\leq$ 1.16) AND (\sepwid $\geq$ 3)\\
        -2.56 & if & (\petwid $\geq$ 1.6)\\ 
        -1.53 & if & (\petlen $\geq$ 5.32) AND (\petwid $\leq$ 1.9)\\ 
        +2.39 & if & (\petlen $\leq$ 4.64) AND (\petlen $\geq$ 3.9) AND (\petwid $\leq$ 1.5) \\
    \addlinespace[1.5ex]
    \multicolumn{3}{l}{\textbf{Neural Rule Ensembles}} \\
        -0.91 & $\times$ & Min( \ReLU(-0.61\halfspace \sepwid-1.06\halfspace \petwid+4.2), \\
        & & \ReLU(1e-4\halfspace \sepwid-2.85\halfspace \petwid +2.8) )\\
        +1.12 & $\times$ & Min( \ReLU(-0.13 \halfspace \sepwid-1.7\halfspace \petwid+3.7), \\
        && \ReLU(-1.46\halfspace \sepwid+1.64\halfspace \petwid+3.7))\\
        -0.69 & $\times$ & Min( \ReLU(-0.86\halfspace \seplen+3.53\halfspace \sepwid-5.3), \\
        && \ReLU(-1.9\halfspace \seplen+1.6\halfspace \sepwid+6.5), \\
        && \ReLU(-0.77\halfspace \seplen+3.5\halfspace \sepwid-5.5) ) \\
        -0.76 & $\times$ & Min( \ReLU(-0.8\halfspace \seplen+3.6\halfspace \sepwid-0.37\halfspace \petlen-4.5), \\
        && \ReLU(0.34\halfspace \seplen+1.5\halfspace \sepwid-0.4\halfspace \petlen-3.6),\\   && \ReLU(-0.87\halfspace \seplen+1.6\halfspace \sepwid-0.97\halfspace \petlen+3.6) )\\
        -0.8 & $\times$ & Min( \ReLU(0.57\halfspace \seplen+3.7\halfspace \sepwid+0.63\halfspace \petlen-17.2), \\
        && \ReLU(1.6\halfspace \seplen+1.15\halfspace \sepwid+0.17\halfspace \petlen-12.2), \\ 
        && \ReLU(0.5\halfspace \seplen+1.5\halfspace \sepwid+0.67\halfspace \petlen-8.6) ) \\
    \addlinespace[1.5ex]
   \multicolumn{3}{l}{\textbf{Explainable Boosting Machine}} \\
    -1.88 & if & True\\
    \multicolumn{3}{c}{
        \includegraphics[width=0.92\linewidth]{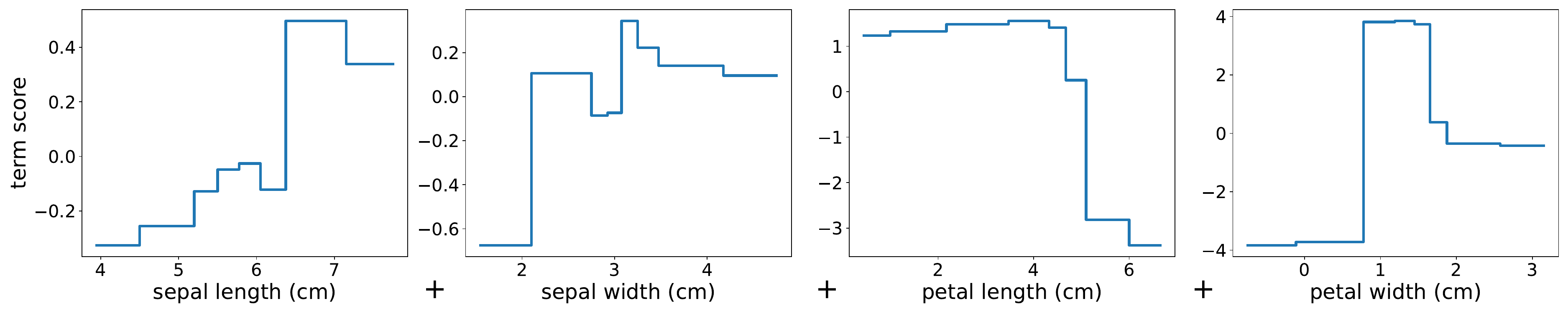}
    }\\
    \bottomrule
    \end{tabular}
    \vspace{-0.7cm}
\end{table}
\section{Additive Rule Ensemble Learning}\label{sec_background}
Probabilistic additive models \eqref{eq:fx} describe the conditional distribution $Y \given X$ through some monotone \defemph{activation function} (also called inverse link function) $\mu$ that, most commonly, translate the value of $f(\bx)$ to the conditional mean of $Y$, i.e., 
\begin{equation*}
\expect(Y \given X=\bx) = \mu(f(\bx)) \enspace .
\end{equation*}
Important examples for this setting include standard regression where $Y \given X$ is assumed to be normally distributed and $\mu(a)=a \in \rev{\bbR}$ is the identi\rev{t}y function, Poisson regression where $Y \given X$ is assumed to be Poisson distributed with $\mu(a)=\exp(a) \in \rev{\bbR_+}$, and logistic regression where $Y \given X$ is assumed to \rev{be} Bernoulli distributed and $\mu(a)=\sigma(a)=1/(1+\exp(-a))\in \rev{(0, 1)}$ is the logistic sigmoid function. 

There are a variety of \defemph{loss functions} $\loss(y,f(\bx))$ that express the goodness of fit of $f$ for a given input / output pair $(\bx, y)$. 
A general construction is to use a \defemph{log-loss} (or deviance function) $-\log P(y\given f(\bx))/a - b$ with positive constants $a$ and $b$ that simplify the computation and ensure that an optimal prediction receives a loss of $0$.
This results, for standard regression, in the \defemph{squared loss} $\loss_{\sqr}(y,f(\bx)) = (y-f(\bx))^2/2$ and for logistic regression in the \defemph{logistic loss} (or cross entropy loss) $\loss_{\log}(y, f(\bx)) = -y\log \rev{\sigma}(f(\bx)) - (1-y)\log(1-\rev{\sigma}(f(\bx)))$.
Importantly, these loss functions are strictly convex in their second arguments.
This is in contrast to the \defemph{0/1-loss}, defined as $\loss_{0/1}(y, f(\bx))= \step(\rev{2}y\rev{-1} \neq \rev{\mathrm{sign}}(f(\bx)))$, which is often considered in the earlier non-probabilistic machine learning literature for classification problems.

The standard supervised learning approach is to learn the model function~\eqref{eq:fx} from training data $(\bx_1, y_1), \dots, (\bx_n, y_n)$ by minimizing the \defemph{regularized empirical risk}
\begin{equation}\label{eq:reg_emp_risk}
    \Risk_\lambda(f) = \lambda\Omega(f)/n + \sum_{i=1}^n \loss(y_i,f(\bx_i)) /n \enspace .
\end{equation}
Here, $\lambda$ is a non-negative \defemph{regularization parameter} that provides a weight to a complexity
measure $\Omega(\cdot)$---typically chosen as the $l1$-norm or the squared $l2$-norm of the coefficient vector, excluding the offset weight $\beta_0$.
That is $\Omega(f)=\|\bbeta\|_1$ or $\Omega(f)=\|\bbeta\|^2_2$ with $\bbeta=(\beta_1, \dots, \beta_r)$.
If the loss function is a log-loss as outlined above, this formulation is equivalent to penalized likelihood maximization. Moreover, in this situation or more generally when the loss function is strictly convex in $f(\bx)$, and when the feature functions $\varphi_1,\dots, \rev{\varphi_r}$ are fixed, minimizing \eqref{eq:reg_emp_risk} just in terms of $\beta_0$ and $\bbeta$ can be solved efficiently through convex optimization.
\rev{For twice differentiable $\ell$}, this can be seen by checking that the Hessian of $\Risk_0$ \rev{as a function of $\beta_0$ and $\bbeta$} is a weighted Gramian matrix $\bPhi \trans \bH \bPhi$ with $\bPhi \in \rev{\bbR^{n \times r+1}}$ being the fixed feature matrix with entries $\phi_{i, j}=\varphi_j(\bx_i)$ \rev{and $\phi_{i,r+1}=1$} and $\bH \in \rev{\bbR^{n \times n}}$ being the diagonal matrix with positive entries $h_{i,i}=\rev{n^{-1}}\partial^2 \loss(y_i, f(\bx_i)) / \partial f(\bx_i)^2$.
\rev{The convexity of $R_\lambda(\beta_0, \bbeta)$ is then implied by the convexity of $\Omega(\bbeta)$ and the fact that a sum of convex functions remains convex.}

To learn the feature functions and the coefficients jointly, there is a range of paradigms that are applicable to rule-ensemble-like structures.
In \defemph{neuro-symbolic learning} approaches\rev{~\citep[e.g.,][]{qiao2021learning, wang2023learning, dierckx2023rl, yu2023finrule, yang2024hyperlogic}}, this is done through \rev{the imposition of} a parametric structure with a fixed number of conjunctions elements $l$ for all feature terms and relaxations of the Boolean features defined by Eq.~\ref{eq:p_poly} to \defemph{soft threshold propositions} $p(\bx; \bw, t)=\sigma(\bw\trans\bx-t)$ or, harder to interpret, \defemph{rectified linear units} $p(\bx; \bw, t) = \max \{\bw\trans\bx - t, 0\}$.
Further, the conjunctions are also sometimes relaxed for computational convenience, e.g., in \defemph{neural rule ensembles} \citep{dawer2020neural}, to $\varphi(\bx; \bW, \bt)=\min \{p(\bx; \bw_1, t_1), \dots, p(\bx; \bw_l, t_l)\}$.
The model learning problem is then amenable to gradient-based methods, which can efficiently produce local minima of the training risk.
Further pre-processing steps can be used to end up with sparse linear transformation vectors.
Neural rule ensembles specifically solves this problem with a non-parametric pre-processing step based on a standard axis-parallel tree model: it makes each feature term correspond to one rule in the tree and only includes learnable non-zero weights for an input variable if the corresponding tree rule uses that variable in one of its propositions.
While computationally convenient, it is important to point out that the resulting rule-like terms are considerably harder to simulate for a human interpreter than the usual conjunctive rules. In particular, their ``rule\rev{s}'' do not have constant outputs, but describe \rev{a} local linear function, the exact value of which depends on the closest of the hyperplane closest to a given test point.
In this work we evaluate neural rule ensembles optimistically by measuring their complexities also with Eq.~\ref{eq:c} as if they were standard rules.

A very different approach to rule ensemble learning is classical \defemph{top-down tree induction}\rev{~\citep[see,][]{breiman2017classification}}, where one starts from a trivial structure $f^{(0)}(\bx) = \beta^{(0)}_1 q^{(0)}_1\!(\bx; \boldsymbol{0}, \mathbf{0})$ with one ``root node'' and then iteratively refines the structure in iteration $i=1, 2, \dots$ by ``splitting'' a Boolean feature, say the one with index $j_i^*$, such that
\begin{equation*}
q^{(i)}_j = 
\begin{cases}
q^{(i-1)}_j &, \text{ if } j < j_i^*\\
q^{(i-1)}_j p(\,\cdot\,; \bw_i^*, t_i^*)&, \text{ if } j = j_i^*\\
q^{(i-1)}_j p(\,\cdot\,; -\bw_i^*, -t_i^*)&, \text{ if } j = j_i^*+1\\
q^{(i-1)}_{j-1} &, \text{ if } j > j_i^* + 1\\
\end{cases} 
\end{equation*}
and stopping either when no such split improves the objective value or some weaker condition is met.
Note that in tree induction all rule conditions are disjoint, hence optimal coefficients can be found for each term easily by one-dimensional convex optimization.
When the splits are restricted to axis-parallel cuts, i.e., $\|\bw\|_0=1$, it is feasible to check all relevant candidate thresholds for all input variables to find the best split in an iteration.
When oblique cuts are allowed, methods for finding candidate cuts include stochastic search~\citep{murthy1994system} or a relaxation of the hard threshold propositions (Eq.~\ref{eq:p_poly}) to soft threshold propositions as defined above, which reduces the problem of oblique split optimization to logistic regression~\citep{irsoy2012soft,carreira2018alternating}.
Importantly, neither of these approaches is guaranteed to find optimal oblique hard threshold propositions and, without modifications, both lead to non-sparse linear transformations that are hard to interpret.

\rev{More generally}, the disjoint rule conditions resulting from tree induction forgo one of the central advantages of rule-based models (which is why they are often not thought of in this framework): the creation of complex prediction functions with a small number of relatively \rev{simple} rules, because creating complex behavior with a small number of rules requires condition overlap.
One way to produce a small ensemble of overlapping rules from a tree induction algorithm is to first generate many alternative tree models in an additive ensemble and then to filter them down, e.g., again by applying $l1$-regularized empirical risk minimization~\citep{friedman2008predictive} or by more complicated extraction procedures~\citep[][]{benard2021interpretable}.
In contrast to other generate-and-select approaches~\citep[e.g.,][]{lakkaraju2016interpretable, wang2017bayesian}, starting with a tree ensemble allows the inclusion of polyhedral conditions through the use of oblique tree learning.
However, it has been demonstrated~\citep{boley2021better,yang2024orthogonal} and we will confirm again here (Secs. \ref{sec_eval} and \ref{sec_accuracy}), that rule ensembles constructed in this way tend to have a substantially worse accuracy/simplicity trade-off than those obtained by applying the ensembling technique of \emph{gradient boosting}~\citep{mason1999functional, friedman2001greedy} directly to \emph{individual} rules.

Here, we focus on the (fully) \defemph{corrective boosting}%
\footnote{The term ``corrective'' refers to performing weight re-computations as  terms are added to the model. As long as the number of terms is small, this can be done with negligible computational cost. In contrast, e.g., \emph{Extreme Gradient Boosting}~\citep[XGBoost][]{chen2016xgboost}, computes weights only once with a fast approximation formula, as it is optimized to create very large additive models. Note that also \emph{Explainable Boosting Machines}~\citep[EBMs,][]{caruana2012intelligible} perform a corrective form of boosting, called \emph{cyclical boosting}---a form of the backfitting algorithm for generalized additive models~\citep{hastie2017generalized}. This is enabled by fixing the terms to one 1d-tree term per variable plus an optional number of 2d-interaction terms.}
variant~\citep{kivinen1999boosting,shalev2010trading, shen2013fully}, which is most suitable to optimize the accuracy/simplicity trade-off. This variant starts with an empty \rev{model} $\rev{f^{(0)}(\bx)}=\beta^{\rev{(0)}}_0 = \argmin(\sum_{i=1}^n \loss(y_i, \beta))$ and for $m = 1,\dots, r$ \rev{iteratively adds terms}
\begin{align}
    f^{(m)}(\bx) &= \beta_0^{(m)} + \beta_{1}^{(m)}q_1(\bx)+\dots+\beta_{m}^{(m)}q_m(\bx) \label{eq:ruleboosting} \\
    \rev{q_m} &\rev{= \argmax \{ \obj(q; f^{(m-1)}) \with q \in \mathcal{Q} \}} \label{eq:qm} \\
     \rev{\beta_0^{(m)}}, \bbeta^{(m)} &= \argmin \rev{\{\Risk_\lambda(f(\,\cdot\, ; q_1,\dots, q_m, \beta_0, \bbeta)) \with \beta_0, \bbeta \in \bbR^{m+1}\} }\label{eq:beta} \enspace .
\end{align}
\rev{Here, $\mathcal{Q}$ denotes the set of available rule conditions, so typically }
\begin{equation}
\rev{\mathcal{Q}_\mathrm{paral}=\{q(\,\cdot\,; \bW, \bt) \with \bt \in \bbR^l, \bW \in \{0, \pm 1\}^{l \times d}, \|\bw_1\|_{0} = \cdots = \|\bw_l\|_0=1\}}
\label{eq:axis_parallel_rule_conditions}
\end{equation}
and $\obj(\,\cdot\,; f)$ is a real-valued \defemph{objective function} that is chosen to approximate the empirical risk reduction achieved by adding a candidate condition \rev{to $f$ (and re-optimizing the weights)}.
\rev{In this work}, we consider the objective function called \defemph{gradient sum} in~\citet{yang2024orthogonal}, which is one of the originally proposed boosting objectives in~\citet{mason1999functional}:
\begin{equation}
    \obj(q; f) = \left| \sum_{i=1}^n q(\bx_i) \frac{\partial \loss(y_i, f(\bx_i))}{\partial f(\bx_i)} \right|
    \label{eq:obj_gs} 
    \enspace .
\end{equation}
Note that the rule weight refitting in \Cref{eq:beta} can again be done efficiently through convex optimization.
To solve \Cref{eq:qm}, the standard approach for axis-parallel cuts, as proposed, e.g., in ~\citet{dembczynski2010ender}, is to start from an empty rule condition $q^{(0)}=1$ and then for $i=1, 2, \dots$ to sequentially apply greedy augmentations
\begin{align}
    q^{(i)} &= q^{(i-1)}p(\,\cdot\,; \bw^{(i)}, t^{(i)})\label{eq:greedy_rule_condition}\\
    \bw^{(i)}, t^{(i)} &= \argmax \{\obj(q^{(i-1)}p(\,\cdot\,; \bw, t)) \with \bw, t \in \rev{\{0,\pm1\}}^d \times \bbR \text{ with }\|\bw\|_0=1\} 
    \label{eq:greedy_w_axis_parallel}
\end{align}
until some maximum complexity or other stopping criterion is reached.
Again, because of the restriction to axis-parallel cuts, \Cref{eq:greedy_w_axis_parallel} can be solved efficiently with exhaustive enumeration.
In the next section we show how the rule boosting approach can be adapted to learn sparse polyhedral conditions.

\section{Sparse Oblique Rule Boosting}\label{sec_method}
In this section, we introduce a base learner for gradient boosting that learns sparse oblique propositions within a rule, as defined in \Cref{eq:q_poly,eq:p_poly}. 
More specifically, the goal is to relax \rev{the restriction of axis parallel cuts in the rule condition set~\eqref{eq:axis_parallel_rule_conditions} to sparse oblique cuts, i.e.,}
\begin{equation*}
\rev{
    \mathcal{Q}_\mathrm{oblique} = \{q(\, \cdot \, ; \bW, \bt) \with \bW \in \bbR^{l \times d}, \zeronorm{\bW} = c_{\max}, \bt \in \bbR^l  \}
}
\end{equation*}
for some \defemph{maximum complexity} hyper-parameter $c_{\max}$.
The proposed base learner solves this problem based on two components: Firstly, a flexible search strategy, described in \Cref{sec:search}, that, motivated by the sparsity requirement, relaxes the classical greedy approach~\eqref{eq:greedy_rule_condition} and that is applicable to all objective functions;
secondly, a reduction of specifically the gradient sum optimization of individual propositions to sparse weighted logistic regression, presented in \Cref{sec:reduction}.

\subsection{Search Strategy}
\label{sec:search}

The straightforward adaption of the greedy rule condition learning given by \eqref{eq:greedy_rule_condition} and \eqref{eq:greedy_w_axis_parallel} would be to replace \eqref{eq:greedy_w_axis_parallel} with the relaxed condition
\begin{equation*}
    \bw^{(i)}, t^{(i)} = \argmax \{\obj(q^{(i-1)}p(\,\cdot\,; \bw, t)) \with \bw, t \in \bbR^{d+1}, \zeronorm{\bw} \leq c_{\max} - \zeronorm{\bW^{(i-1)}}\} 
\end{equation*}
where $\bW^{(i)}$ denotes the weight matrix resulting from stacking the learned row vectors $\bw^{(1)}, \dots, \bw^{(i)}$.
While this strategy would allow for oblique cuts to be learned,
it has the fundamental flaw of encouraging dense linear transformations and is wasteful in the allocation of the overall complexity budget. 
In any given greedy step this simplistic approach considers first all oblique cuts before adding another proposition.
Thus it fails to identify combinations of simple individual propositions.
This is exacerbated by potentially overfitting noise in the input variables.

\begin{figure}[t]
    \centering
    \includegraphics[width=1.08\textwidth]{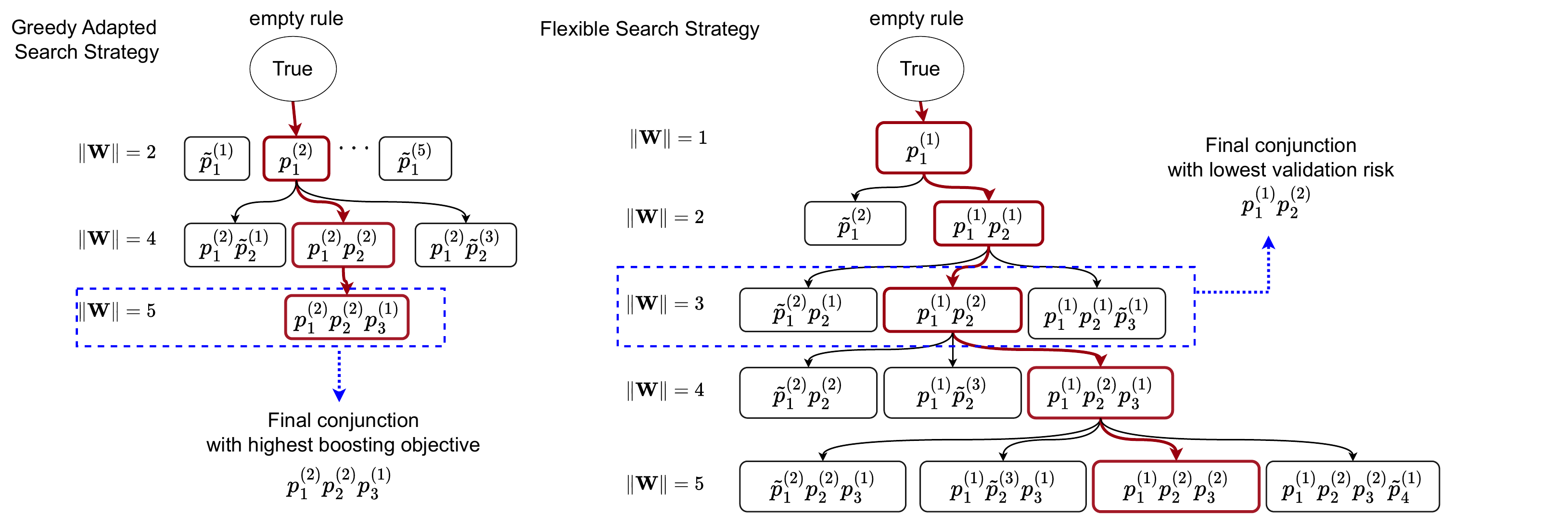}
    \caption{Comparison of two search strategies for learning a single conjunction with a maximum rule complexity of $c_{max}=5$. Each box represents a candidate conjunction including candidate proposition $\tilde{p}_j^{(s)}$ with sparsity $\zeronorm{\bw_j}=s$, while $p_j^{(s)}$ denotes a fixed selected proposition. The red paths indicate the best conjunction candidates selected at each complexity level. In both strategies, a new proposition can be added at each level to increase rule complexity.
    \textbf{(Left)}: At each complexity level (vertical levels), all possible sparsity levels (horizontal levels) for the next proposition while previously learned propositions are fixed. The conjunction candidate with the highest boosting objective is selected as the final conjunction. 
    \textbf{(Right)}: At each complexity level, the conjunction complexity increases by one, allowing previously learned propositions to be refitted with higher sparsity ($s+1$). All configurations constrained by the current complexity are evaluated via the boosting objective, and the best candidate is retained. Finally, selected conjunctions across all complexity levels are compared on validation risk, and the one with the lowest risk is selected as the final rule.}
    \label{fig:search_strategy}
\end{figure}

Therefore, we employ two measures to refine this strategy: (i) instead \rev{of} increasing the complexity of the weight matrix in arbitrary steps, we systematically enumerate candidate solutions for all complexity levels from $1$ to $c_{\max}$ and (ii) instead of selecting the candidate weight matrix with the highest boosting objective, we choose the one that leads to the lowest validation risk after re-fitting \eqref{eq:beta}. See \Cref{fig:search_strategy} for an intuitive comparison of this strategy to the one that simply adapts the traditional greedy search for axis-parallel conditions.  

In more detail, the method obtains one candidate matrix $\bW^{(i)} \in \bbR^{l_i \times d}$ for each complexity level $i=1,\dots, c_{\max}$ by starting with 0-row matrix $\bW^{(0)}$, and then, for $i=1,\dots,c_{\max}$ to set $\bW^{(i)}$ to the optimal matrix among a set of ``refinements'' of $\bW^{(i-1)}$, each of which corresponds to a complexity increase by one.
Specifically, we obtain one refinement $\bW^{(i, j)}$ for each  proposition / row $j = 1\dots, l_{i-1}$ of $\bW^{(i-1)}$ plus an additional refinement $\bW^{(i, l_{i-1}+1)}$ that adds one new proposition.
The refinement corresponding to an existing row $\bw^{(i-1)}_l$, is defined as the optimal weight vector with complexity increased by one, where all components including the selected variables, their coefficients, and the threshold are allowed to change relative to the previous complexity level. The refinement that adds \rev{a} new proposition contains the best augmentation with a new proposition of complexity $1$ (similar to the classical search scheme).
Formally, the refinements for $j = 1,\dots,l_{i-1}+1$ are defined as
\begin{align}
    \bW^{(i,j)}, \bt^{(i,j)} &= \argmax \{ \obj(q(\cdot; \bW, \bt);f^{(m-1)}) \with \bW \in \rev{\bbR^{(l_{i-1}+1)\times d}}, \bt \in \rev{\bbR^{l_{i-1}+1}},\label{eq:poly_search}\\ 
    & \hspace{0.5cm} \zeronorm{\bw_j} = \zeronorm{\bw_j^{(i-1)}} + 1 ,\bw_l = \bw^{(i-1)}_l \text{ for } l \in \{1, \dots, l_{i-1}+1\} \setminus \{j\} \} \nonumber 
\end{align}
where we define $\bw_{l_{i-1}+1}^{(l-1)} \in \bbR^d$ to be an all-zero vector.
This convention ensures that the new proposition introduced by the last refinement is always a complexity $1$ proposition, as desired.
\begin{align}
    j_* &= \argmax \{ \obj(f^{(m-1)}q(\cdot; \bW^{(i, j)}, \bt^{(i, j)})) \with j=1, \dots, l_{i-1}+1\} \\
    \bW^{(i)} &= \bW^{(i, j_*)}
\end{align}
We can see that with \rev{this} scheme, indeed there is one candidate per complexity level, because $\zeronorm{\bW^{(i)}} = \zeronorm{\bW^{(i-1)}}+1$.
It remains to choose, which of those candidates to select across complexity levels $i = 1, \dots, c_{\max}$.

To do so, we evaluate the ensemble response of each candidate $q(\,\cdot\,; \bW^{(i)}, \bt^{(i)})$ on a held-out validation set and choose the one that achieves the lowest validation risk
\begin{align}
      \bW_m &= \bW^{(i_*)} \nonumber\\
 i_* &= \argmin\{\Risk^{\text{val}}(\by,f^{(m-1)}(\,\cdot\,) q(\,\cdot\,; \bW^{(i)},\bt^{(i)}))) \with i = 1,\dots,c_{\max}\} \enspace.
\end{align}
With the overall search strategy in place, it remains to design a method for solving \Cref{eq:poly_search}, i.e., to learn an optimal proposition of prescribed sparsity, with all other propositions fixed.
While the proposed strategy is applicable to all boosting objectives, in the next section, we develop an approach for solving \Cref{eq:poly_search} specifically for the gradient sum objective (\Cref{eq:obj_gs}).
\subsection{Logistic Linear Transformation Boosting}\label{sec:reduction}
The key insight for maximizing the gradient sum objective for oblique hyperplane conditions, is that finding optimal weights can be reduced to a weighted linear classification problem based on the 0/1-loss.
As it has been described in \Cref{eq:obj_gs}, gradient sum objective is a function of rule conditions and gradients of the loss function with respect to the ensemble response. In this context, for $i=1,\dots,n$, let $g(\bx_i) = \partial \loss(y_i, f^{(m-1)}(\bx_i))/\partial f^{(m-1)}(\bx_i)$, $\bg = (g(\bx_1),\dots, g(\bx_n))$ and $\bq = (q(\bx_1),\dots, q(\bx_n))$.
We firstly break down the absolute value in~\eqref{eq:obj_gs} into positive and negative parts, $\obj_+(\bq)=\inner{\bg}{\bq}$ and $\obj_-(\bq)=\inner{-\bg}{\bq}$, and write 
\begin{equation}
\max |\inner{\bg}{\bq}|=\max \{\inner{\bg}{\bq}, \inner{-\bg}{\bq} \! : \, \bq \in \setQ_{\rev{\mathrm{oblique}}}\} \enspace .
\end{equation}
Then, given the set of still selectable training examples to learn the $j$-th proposition of an individual $q$, $\setI_{j}=\{i \with p_1(\bx_i)=\dots=p_{j-1}(\bx_i)=p_{j+1}(\bx_i)=p_l(\bx_i)=1\}$, we rewrite the objective for $\signg \in \{-1, 1\}$ as follows:
\begin{align}
    \obj^{(\signg)}_{j}(\bw, t) &= \sum_{i\in \setI_j} \signg g_i \step\{\bx_i\trans\bw \geq t\} \\
    &=\sum_{i\in \setI^+_j} \signg g_i - \sum_{i\in \setI^+_j} \signg g_i\step\{\bx_i\trans\bw < t\} + \sum_{i\in \setI^-_j} \signg g_i\step\{\bx_i\trans\bw \geq t\}\\
    &= \sum_{i\in \setI^+_j} \signg g_i - \underbrace{\sum_{i \in \setI_j} |g_i| \loss_{01}(\signg \mathrm{sign}(g_i), \step\{\signg \bx_i\trans\bw\geq t\})}_{\Risk^{(\signg)}_{01}(\bw, t)} \label{eq:o_row3}
\end{align}
where we used the symbols $\setI^+_j = \{i \in \setI_j \with \signg g_i \geq 0 \}$ and $\setI^-_j = \{i \in \setI_j \with \signg g_i < 0 \}$.
Thus, maximizing the objective function for proposition $j$ is equivalent to minimizing the weighted sum of 0/1-losses when interpreting the weight vector as a linear separating hyperplane for predicting the gradient signs.
This allows us to utilize linear classification algorithms to optimize the weight parameters.
Before we do that, we incorporate the idea of controlling the overall complexity of an individual rule at complexity level, $c$, meaning $\zeronorm{\bW}=c$, into the process of learning the proposition parameters. To find the optimal $\bw^*_j$ and $t^*_j$ at a given sparsity level $s$, where $s\leq c$ to maintain $\zeronorm{\bW}=c$, we solve a constrained optimization problem that enforces this sparsity during the learning of each proposition.
\begin{equation}
    \Risk^{(\signg)}_{01}(\bw^*_j,t^*_j) = \min \{\Risk^{(\signg)}_{01}(\bw,t) : t \in \bbR, \bw\in \bbR^d, \zeronorm{\bw}=s\} \enspace .
\end{equation}
Since this is a hard optimization problem, it is commonly optimized using a convex surrogate function. In this work, we propose to optimize the $l1$-regularized risk with respect to the logistic loss, $\loss_{\log}$, i.e., to find parameters
\begin{equation}\label{eq:l1logreg}
    \bw^*_j,t^*_j = \argmin \Risk^{(\signg)}_{\log}(\bw,t; |\bg|) + \lambda_s\onenorm{\bw}
\end{equation}
where $\lambda_s$ is a regularization value that results in a solution with exactly $s$ non-zero weights and $\Risk^{(\signg)}_{\log}$ refers to the same sub-expression of \Cref{eq:o_row3} as $\Risk^{(\signg)}_{01}$ just with $\ell_{01}$ replaced by $\ell_{\log}$. We identify this parameter using a bisection search strategy while solving the convex optimization problem in \Cref{eq:l1logreg}. Once the relevant features are selected, we refit the coefficients $\bw^*$ and threshold $t^*$ using an unregularized logistic regression, restricted to the selected features. This refitting step reduces the estimation bias introduced by the initial $l1$-regularization. In this work,  we use ``LibLinear"~\citep{fan2008liblinear} as an efficient solver for~\eqref{eq:l1logreg}.
\subsection{Computational Complexity}
To analyze the asymptotic computational complexity of the proposed method in terms of the number of rules $r$, the dimensionality $d$, and amount of training data $n$, as well as the maximum complexity level per rule condition $c_{\max}$, we start with the cost of learning the rule conditions, specifically with the key step of solving \Cref{eq:l1logreg}.
For a fixed $\lambda$, assuming $d\leq n$, this is done in time $O(d^2 n S_1)$ by a Newton-like iterative solver such as iteratively re-weighted least squares (\rev{IRLS}).
This corresponds to the cost of computing the weighted Gramian $\bX\trans\bLambda_i\bX$ for iterations $i=1, \dots, S_1$ where $S_1$ is the number of iterations required to reach the desired convergence criterion.
Note that, while this number does not directly depend on $d$ or $n$, we leave it explicit in the analysis as it constitutes a cost factor unique to the proposed method.
To find a regularization value $\lambda_s$ that results in an $s$-sparse solution, the iterative solver is then called $B \in O(\log \lambda_{\max} / (b_s-a_s))$ times within a bisection search.
While this number is hard to analyze theoretically, an empirical investigation (see \Cref{sec_bisection_search}) shows that it can be largely treated as a constant (less than 20 across all datasets) with  perhaps only mild increasing dependency on $d$.

Further, the search procedure applies bisection search to find candidate rule conditions for each complexity level, $1\dots, c_{\max}$.
In the worst case, the number of propositions at each complexity level is maximal, i.e., when determining the complexity level $c+1$ candidate, the complexity level $c$ candidate to be refined consists of $c$ axis-parallel propositions, resulting in $c+1$ refinements to be checked.
Overall, this results in $1 + 2 + \dots + c_{\max} \in O(c^2_{\max})$ calls to the bisection search.
This leaves us with a total time complexity of $O(rc^2_{\max} d^2n BS_1)$ for all of the $r$ rule condition optimization steps.
Here we note that, while the number of relevant data points in a given logistic expression problem $|\setI_j|$ can be less than $n$, in the worst case it is still order of $n$ (when each proposition only deselects a constant number of data points).
Finally, the cost of refitting the rule weights is not different for the proposed method than for any other (fully) corrective boosting algorithm. In contrast to the condition learning, the cost of this part of boosting iteration $k$ does depend on $k$ through the growing feature matrix. 
While incremental least squares can be applied to solve the problem more efficiently for regression, in general we have to apply an iterative algorithm like IRWLS to the current feature matrix $\bQ_k \in \bbR^{n \times k}$ with entries $q_{i,j}=q(\bx_i, \bW_j, \bt_j)$.
For $r \leq n$, which we can safely assume for interpretable modeling,
this operation is again dominated by the computation of the weighted Gramian $\bQ_k\trans\bLambda\bQ_k$, resulting in a complexity of $O(k^2nS_2)$ where $S_2$ is the number of iterations required by the IRLS procedure to reach the desired convergence criterion. This number does not directly depend on $d$, $n$, or $r$ (though it can increase in $r$ when many similar rules are added).
Adding up this cost across all boosting iterations results in a total cost in $O(r^3nS_2)$ for the weight refitting step in general (which can be reduced by a factor of $r$ for the standard regression model).
In summary, the total worst case time complexity can be bounded as
\begin{equation}\label{eq:comp_complexity}
    T_{\mathrm{LLTB}} \in O(rn(\underbrace{c_{\max}^2 d^2 BS_1}_{\text{learn } \bW, \bt} + \underbrace{r^2S_2}_{\text{refit }\bbeta})) \enspace .
\end{equation}
It can be seen that, for small numbers of rules, the cost of the full weight refitting is negligible relative to the cost of rule condition learning.
Importantly, the time complexity is strictly linear in the number of training data points $n$---in contrast to methods that rely on presorting the data points for fast incremental axis-parallel cut point search. 
\begin{figure}[ht!]
    \centering
    \includegraphics[width=1\textwidth]{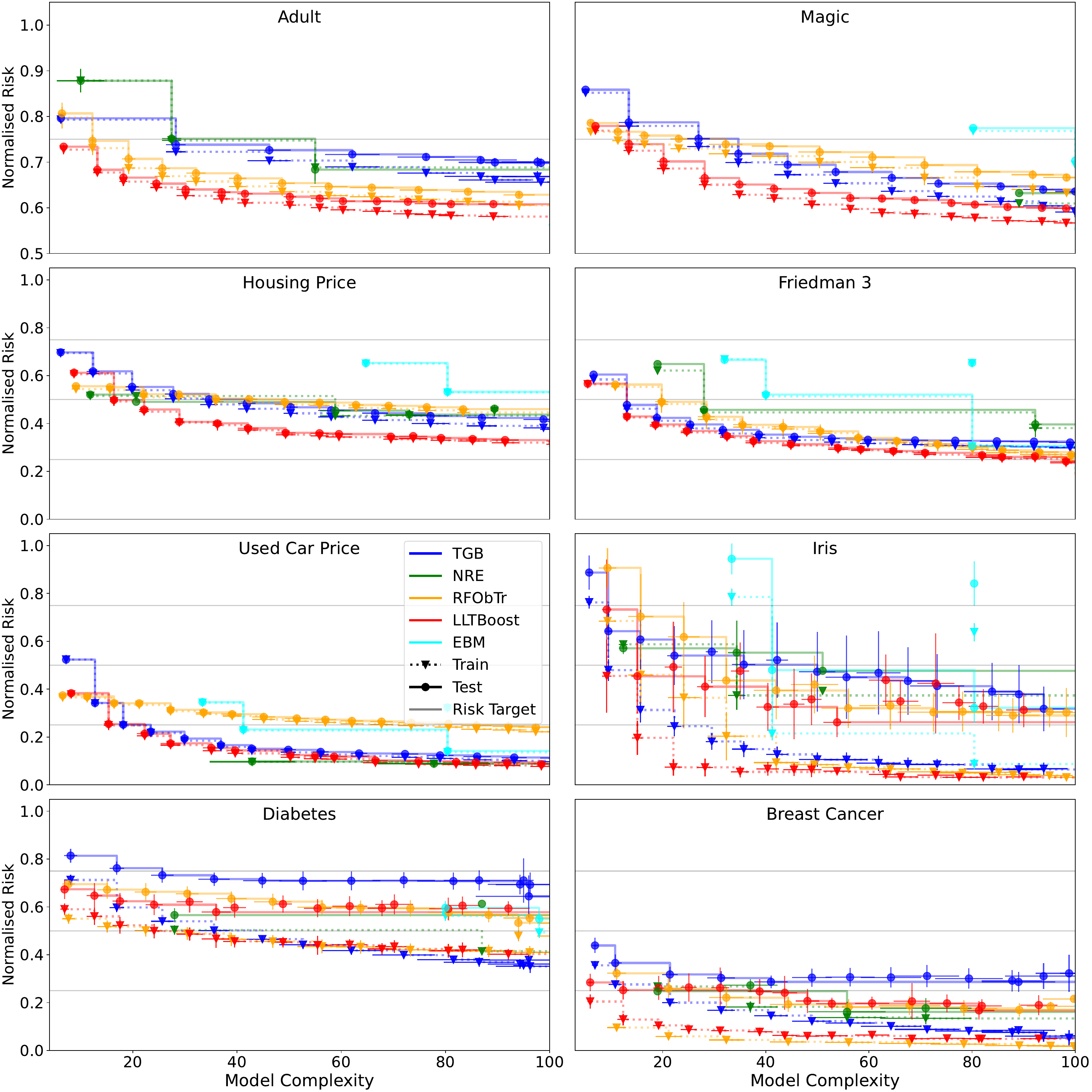}
    \vspace{-0.2cm}
    \caption{Risk/complexity trade-offs for different methods and 8 datasets based on \textbf{cross-validation-selected} hyper-parameters (corresponding plots for the remaining datasets are provided in appendix, \Cref{fig:complexity_risk_cv_trends_append}). The y-axis shows the mean normalized risk, computed using squared loss for regression datasets and logistic loss for classification datasets, averaged over 15 repetitions for model complexities ranging from 1 to 100. Error bars represent 95\% confidence intervals.}
    \label{fig:complexity_risk_cv_trends}
\end{figure}
\section{Empirical Evaluation}\label{sec_eval}
In this section, we compare the risk/complexity trade-off of LLTBoost to that of other rule learning methods across a selection of 14 benchmark \href{https://github.com/fyan102/FCOGB/tree/main/datasets}{datasets} from \citet{yang2024orthogonal}, focusing on classification and least-squares regression datasets with primarily numerical features.
\subsection{Methods}
In the primary study presented here\footnote{The appendix presents additional experiments with an additional baseline, Bayesian Rule Ensembles (\Cref{sec_accuracy}), as well as idealized ``oracle'' versions of all methods with $l2$-regularization (\Cref{sec_oracle_results}).}, we include the following methods (see \Cref{table:iris_rules,table:housing_rules} for illustrations of \rev{the rules produced by these methods}):
\begin{itemize}
    \item \textbf{Logistic Linear Transformation Learning (LLTB)} This is the proposed method, the Python implementation of which is publicly available on GitHub\footnote{\href{https://github.com/shahrzaddii/obliqueruleboosting.git}{https://github.com/shahrzaddii/obliqueruleboosting.git}}. The maximum complexity per rule is set to $c_{\max} = 5$. 
    The $l2$-regularization parameter is chosen by internal $5$-fold cross validation on the training set with subsequent model refitting using all training data.
    Different risk/complexity trade-offs are realized by varying the number of rules $r \in \{1, 2, \dots \}$.

    \item \textbf{Traditional Gradient Boosting (TGB)} This baseline is the classical additive rule-learning framework with gradient boosting of axis-parallel rules (as defined in \Cref{sec_background,eq:ruleboosting,eq:qm,eq:beta}) using the implementation of~\citet{yang2024orthogonal}\footnote{\href{https://github.com/fyan102/FCOGB}{https://github.com/fyan102/FCOGB}}. 
    In this implementation, each rule is constructed greedily by adding propositions one at a time, searching over all feature–threshold pairs to maximize the boosting objective. 
    The $l2$-hyper-parameter is treated in the same way as done for LLTB.
    Also as for LLTB, the number of rules is used to realize different risk/complexity trade-offs.
    
    \item \textbf{RuleFit with Oblique Trees (RFObTr)} As another approach to learn small oblique rule ensembles as defined in \Cref{eq:fx_poly,eq:q_poly,eq:p_poly}, this baseline combines the RuleFit generate-and-select method~\citep{friedman2008predictive} based on $l1$-regularized risk minimization with bagged forests of oblique classification and regression trees. 
    Specifically, we use the publicly available Python package \texttt{obliquetree}\footnote{\href{https://github.com/sametcopur/obliquetree/tree/main}{https://github.com/sametcopur/obliquetree/tree/main}} to fit individual oblique trees, which implements a blend of the approaches mentioned in Sec.~\ref{sec_background}.
    With this implementation, oblique candidate splits are formed by exhaustively searching over the $d$ choose $s_{\mathrm{ob}}$ variable subsets of size $s_{\mathrm{ob}}$.
    Each forest contains $40$ trees using combinations of $\mathrm{max\_depth} \in \{2,3\}$ and $s_{\mathrm{ob}} \in \{2,3,4,5\}$.
    Further, to produce results as consistent as possible with the previous baselines, binary search is used to determine the $\lambda_{l1}$ that selects the desired number of rules $r \in \{1, 2, 3, \dots\}$, and final rule weights are found with $l2$-regularized empirical risk minimization where $\lambda_{l2}$ is determined in the same way as for LLTB and TGB.

    \item \textbf{Neural Rule Ensembles (NRE)} This method, introduced in~\citet{dawer2020neural}, represents the neuro-symbolic learning approach that combines rules extracted from axis-parallel classification and regression trees with joint continuous optimization of all model parameters. 
    As discussed in \cref{sec_background}, the model architecture replaces conjunctive combinations of oblique cuts with MIN-pooling of ReLU functions.
    Generated rules generated have a different structure as illustrated in \Cref{table:iris_rules}. While the original method introduces a margin-based splitting criterion for classification, we adapt it for regression by using mean squared error for tree training. To generate different risk/complexity trade-offs, we vary the maximum tree depth parameter $\mathrm{max\_depth} \in \{2,4,6\}$.

    \item \textbf{Explainable Boosting Machine (EBM)} This approach, proposed in ~\citet{lou2013accurate}, models $f(\bx)$ as a sum of $d$ univariate step functions---one per input variable---with a user-selected number of $b$ bins and a user selected number $\kappa$ of bi-variate interactions terms, which are again piecewise constant functions.
    This can be seen as an instance of the rule ensemble model given by \Cref{eq:q_poly,eq:p_poly,eq:fx_poly} with one rule per constant piece (interval or rectangle).
    Specifically, there are $r=db+ \kappa b^2$ many rules, corresponding to the univariate and the interaction terms, respectively.
    This means that the rules can be grouped by the variable or variable combination that they refer to, and there are $d$ groups of size $b$ and $\kappa$ groups of size $b^2$.
    Importantly, thresholds are shared between consecutive members of these groups.
    This constraint means that one can mentally store this model variant more efficiently than without it---with implications for the complexity metric discussed below.
    Here, we use the implementation of the \texttt{InterpretML} package~\citep{nori2019interpretml} and set $\kappa \in \{1, \dots, 5\}$ and $b \in \{4, 5, 10\}$ to generate different risk/complexity trade-offs ($b=3$ was tested but found to be an invalid).
    The method also relies on $l2$-regularization for fitting the rule weights, which is handled as for the other baselines.

\end{itemize} 
\subsection{Protocol}
All methods can be configured to realize a range of different risk/complexity trade-offs.
It is thus necessary to compare their risk/complexity \emph{curve} (or Pareto front). 
For that purpose, the methods are run with hyper-parameters corresponding to increasing model complexities until the complexity exceeds 100.
Here the model complexities are defined as in \Cref{eq:c} for LLTB, TGB, RFObTr, and NRE.
For NRE this constitutes an optimistic estimate of the actual complexity, because, the rule form with MIN-pooled ReLU functions means that rule outputs depend on the distance to the closest oblique hyperplane to a given test point.
That means that, when simulating NRE model outputs in the same way as the other models, one actually needs to decompose the additive terms further, corresponding to the sub-areas related the individual hyperplanes (in addition to the difficulty of computing Euclidean distances).
For EBM, a straightforward application of the measure \Cref{eq:c} would result in a complexity of $5r$, but that would disregard the reduced storage requirement for memorization due to the systematic relation of the individual rules (of consecutive intervals / rectangles). Therefore, for EBM models, we only account for a complexity of $2r$ corresponding to the different output values and a single storage of the shared threshold vector.
For each complexity level $c$ we then consider the normalized test risk 
\begin{equation} \label{eq:normrisk}
\Risk_{\mathrm{norm}}(c) = \Risk_{\mathrm{test}}(c) / \Risk(0),
\end{equation}
where $\Risk_{\mathrm{test}}(c)$ is the test risk of the trained model (as usual based on squared loss for regression and logistic loss for classification), and $\Risk(0)$ is the test risk of an ``empty'' model with only an offset / background rule that is fit to the training data.
This means that in case of regression the risk values are normalized by the label variance and for classification by the label entropy.
In this way, we obtain a goodness-of-fit metric that is comparable across datasets with different label distributions and prediction tasks.
All methods are run over 15 repetitions of random train/test splits generated via bootstrap sampling for training and the corresponding out-of-bag samples for testing. The bootstrap training size is set to $n' = \min(n, 3000)$ to limit computational cost.
\subsection{Results}
Figures~\ref{fig:complexity_risk_cv_trends} and \ref{fig:complexity_risk_cv_trends_append} illustrate the trade-off between model complexity and normalized risk of all $14$ datasets in terms of the mean $c$ and $\Risk(c)$ values across the 15 repetitions. 
The Pareto fronts are connected by straight lines and error bars in both dimensions are added based on standard 95\% confidence intervals around the mean.
We can see that LLTBoost achieves the lowest overall test risk on the majority of datasets within the complexity limit of 100.
Note that none of the EBM model complexities for \texttt{adult} and \texttt{breast cancer} fall within this range, as they exceed the considered complexity budget.
Additionally, \Cref{table:risk_cv} presents a comparison of normalized test risks across all datasets, averaged over all risk/complexity trade-offs up to complexity 100. Again, the proposed method has the lowest such average on a majority of dataset.

Further, we can ``scan'' the Pareto front of each method by comparing the minimum complexity required to each certain target risk levels.
Specifically, for each method, we identify the ensemble with the lowest complexity $c$ that achieves the normalized test risk reduction targets $\tau \in \{0.25, 0.5, 0.75\}$, i.e., $\Risk_{\mathrm{norm}}(c)\leq 1-\tau$.
If none of the ensembles within the complexity budget achieve the target, i.e, $\Risk_{\mathrm{norm}} > 1-\tau$ for all ensembles, we pessimistically assume that the model cannot reach the risk target at all and report the complexity as infinity ($\infty$).
This choice implies that standard mean aggregation over repetitions is unsuitable, because one individual repetition where a risk target is not met would cause the mean complexity to become infinity.
We therefore resort to reporting median required complexities and, correspondingly, use non-parametric confidence intervals of the median given by the $5$-th and $11$-th order statistics, which corresponds to $92\%$ confidence intervals (see~\Cref{sec_median_ci} and~\cite{hahn2011statistical} for details).
\Cref{table:complexity_cv} summarizes the resulting model complexities for each method under three levels of normalized test risk target reduction.
Based on these results, LLTBoost achieves the lowest model complexity on the majority of datasets across all three risk reduction targets, especially for higher accuracy with 75\% risk reduction. \rev{This is confirmed by a complementary evaluation of the risks using a target model complexity budget of up to $50$ presented in \Cref{table:risk_based_on_target_complexity}.}

Finally, \Cref{table:comp_time} reports the computation time corresponding to the median complexity reported in \Cref{table:complexity_cv} with 25\% risk reduction. 
As it can be seen, LLTBoost method does not lead to a significant computation overhead compared to TGB. It is worth noting that the computational cost of RFObTr is particularly sensitive to the number of trees, depth of trees, and number of features used in each linear split. Similarly, the runtime of EBM depends strongly on the number of bins per feature or interaction term.
\begin{table}
\caption{Model complexity comparison of methods across benchmark datasets, for three normalized test-risk reduction targets, based on \textbf{cross-validation-selected} hyper-parameters. For each target, the first ensemble with the smallest complexity that hits the target is selected. Three numbers are
reported for each experiment, representing the median (center), lower (left), and
upper (right) bounds of the 92\% confidence interval.}\label{table:complexity_cv}
\centering
\begin{tabular*}{\textwidth}{@{\extracolsep\fill}lccccc}
\toprule%
& \multicolumn{5}{c}{25\% risk reduction} \\ \cmidrule{2-6}
Dataset & TGB	& NRE & RFObTr & LLTB & EBM\\
\midrule
Iris & 
\lomedup{6}{\textbf{8}}{13} & \lomedup{10}{10}{13} & \lomedup{16}{25}{29} & \lomedup{8}{9}{14} & \lomedup{40}{40}{40}\\
Liver & \lomedup{\inft}{\inft}{\inft} & \lomedup{\inft}{\inft}{\inft} & \lomedup{\inft}{\inft}{\inft} & \lomedup{\inft}{\inft}{\inft} &  \lomedup{\inft}{\inft}{\inft}\\
Diabetes & 
\lomedup{14}{16}{37} & \lomedup{28}{28}{28} & \lomedup{6}{\textbf{7}}{9} & \lomedup{7}{\textbf{7}}{8} & \lomedup{80}{80}{80}\\
Breast Cancer & 
\lomedup{7}{7}{7} & \lomedup{36}{50}{66} & \lomedup{7}{13}{13} & \lomedup{6}{\textbf{6}}{7} &\lomedup{\inft}{\inft}{\inft} \\
Banknote & \lomedup{3}{\textbf{3}}{3} & \lomedup{48}{50}{54} & \lomedup{9}{9}{11} & \lomedup{5}{5}{6} & \lomedup{32}{32}{32}\\
Red Wine &
\lomedup{25}{38}{55} & \lomedup{28}{28}{28} & \lomedup{11}{13}{19} & \lomedup{11}{\textbf{12}}{14}  & \lomedup{88}{88}{88}\\
Car Price & \lomedup{5}{5}{5} & \lomedup{28}{28}{28} & \lomedup{4}{\textbf{4}}{5} & \lomedup{5}{6}{8} & \lomedup{32}{32}{32}\\
Friedman1 & \lomedup{10}{12}{12} & \lomedup{28}{28}{28} & \lomedup{7}{7}{7} & \lomedup{6}{\textbf{6}}{7} &\lomedup{80}{80}{80}\\
Friedman2 &
\lomedup{5}{\textbf{5}}{5} & \lomedup{19}{19}{19} & \lomedup{4}{\textbf{5}}{5} & \lomedup{5}{7}{8} & \lomedup{32}{32}{32}\\
Friedman3 & \lomedup{7}{7}{7} & \lomedup{28}{28}{28} & \lomedup{11}{11}{11} & \lomedup{5}{\textbf{5}}{6} & \lomedup{32}{32}{32}\\
Magic & \lomedup{27}{29}{36} & \lomedup{\inft}{\inft}{\inft} & \lomedup{21}{27}{44} & \lomedup{12}{\textbf{13}}{17} & \lomedup{100}{100}{100}\\
Adult & \lomedup{26}{28}{30} & \lomedup{34}{\inft}{\inft} & \lomedup{12}{13}{17} & \lomedup{7}{\textbf{7}}{7} & \lomedup{\inft}{\inft}{\inft}\\
Voice & \lomedup{3}{\textbf{3}}{3} & \lomedup{51}{69}{82} & \lomedup{7}{9}{13} & \lomedup{7}{7}{7} & \lomedup{\inft}{\inft}{\inft}\\
Housing Price & \lomedup{3}{\textbf{5}}{5} & \lomedup{10}{10}{10} & \lomedup{7}{7}{7} & \lomedup{7}{7}{8} & \lomedup{64}{64}{64}\\
\botrule
\textbf{\# Wins} & 5 & 0 & 3 & 7 & 0\\
\midrule
& \multicolumn{5}{c}{50\% risk reduction} \\ \cmidrule{2-6}
Iris & \lomedup{13}{24}{39} & \lomedup{\inft}{\inft}{\inft} & \lomedup{20}{27}{34} & \lomedup{9}{\textbf{12}}{16} & \lomedup{40}{40}{40}\\
Liver & \lomedup{\inft}{\inft}{\inft} & \lomedup{\inft}{\inft}{\inft} & \lomedup{\inft}{\inft}{\inft} & \lomedup{\inft}{\inft}{\inft} & \lomedup{\inft}{\inft}{\inft}\\
Diabetes & 
\lomedup{\inft}{\inft}{\inft} & \lomedup{\inft}{\inft}{\inft} & \lomedup{\inft}{\inft}{\inft} & \lomedup{\inft}{\inft}{\inft} & \lomedup{\inft}{\inft}{\inft}\\
Breast Cancer & 
\lomedup{7}{7}{9} & \lomedup{36}{50}{66} & \lomedup{9}{13}{13} & \lomedup{6}{\textbf{6}}{7} & \lomedup{\inft}{\inft}{\inft} \\
Banknote & \lomedup{10}{10}{10} & \lomedup{48}{50}{54} & \lomedup{9}{9}{11} & \lomedup{5}{\textbf{5}}{6} & \lomedup{32}{32}{32}\\
Red Wine &
\lomedup{\inft}{\inft}{\inft} & \lomedup{\inft}{\inft}{\inft} & \lomedup{\inft}{\inft}{\inft} & \lomedup{\inft}{\inft}{\inft}  & \lomedup{\inft}{\inft}{\inft}\\
Car Price & \lomedup{5}{10}{12} & \lomedup{28}{28}{28} & \lomedup{4}{\textbf{4}}{5} & \lomedup{5}{6}{8} & \lomedup{32}{32}{32}\\
Friedman1 & \lomedup{25}{26}{28} & \lomedup{28}{28}{\inft} & \lomedup{7}{\textbf{7}}{7} & \lomedup{6}{13}{15} & \lomedup{80}{80}{80}\\
Friedman2 &
\lomedup{5}{\textbf{5}}{5} & \lomedup{19}{19}{19} & \lomedup{4}{\textbf{5}}{5} & \lomedup{5}{7}{8} & \lomedup{32}{32}{32}\\
Friedman3 & \lomedup{14}{14}{14} & \lomedup{28}{28}{28} & \lomedup{18}{28}{33} & \lomedup{12}{\textbf{13}}{14} & \lomedup{80}{80}{80}\\
Magic & \lomedup{\inft}{\inft}{\inft} & \lomedup{\inft}{\inft}{\inft} & \lomedup{\inft}{\inft}{\inft} & \lomedup{\inft}{\inft}{\inft} & \lomedup{\inft}{\inft}{\inft}\\
Adult & \lomedup{\inft}{\inft}{\inft} & \lomedup{\inft}{\inft}{\inft} & \lomedup{\inft}{\inft}{\inft} & \lomedup{\inft}{\inft}{\inft} & \lomedup{\inft}{\inft}{\inft}\\
Voice & \lomedup{3}{\textbf{3}}{3} & \lomedup{51}{69}{82} & \lomedup{7}{9}{13} & \lomedup{7}{7}{7} & \lomedup{\inft}{\inft}{\inft}\\
Housing Price & \lomedup{37}{38}{44} & \lomedup{\inft}{\inft}{\inft} & \lomedup{32}{35}{45} & \lomedup{16}{\textbf{20}}{23} & \lomedup{\inft}{\inft}{\inft}\\
\botrule
\textbf{\# Wins} & 2 & 0 & 3 & 5 & 0\\
\midrule
& \multicolumn{5}{c}{75\% risk reduction} \\ \cmidrule{2-6}
Iris & 
\lomedup{50}{89}{\inft} & \lomedup{\inft}{\inft}{\inft} & \lomedup{65}{76}{\inft} & \lomedup{20}{\textbf{32}}{56} & \lomedup{80}{\inft}{\inft}\\
Liver & \lomedup{\inft}{\inft}{\inft} & \lomedup{\inft}{\inft}{\inft} & \lomedup{\inft}{\inft}{\inft} & \lomedup{\inft}{\inft}{\inft} & \lomedup{\inft}{\inft}{\inft}\\
Diabetes & 
\lomedup{\inft}{\inft}{\inft} & \lomedup{\inft}{\inft}{\inft} & \lomedup{\inft}{\inft}{\inft} & \lomedup{\inft}{\inft}{\inft} & \lomedup{\inft}{\inft}{\inft}\\
Breast Cancer & 
\lomedup{92}{\inft}{\inft} & \lomedup{63}{66}{\inft} & \lomedup{13}{33}{42} & \lomedup{7}{\textbf{9}}{21} & \lomedup{\inft}{\inft}{\inft}\\
Banknote & \lomedup{33}{35}{42} & \lomedup{\inft}{\inft}{\inft} & \lomedup{9}{9}{11} & \lomedup{5}{\textbf{5}}{6} & \lomedup{40}{40}{40}\\
Red Wine &
\lomedup{\inft}{\inft}{\inft} & \lomedup{\inft}{\inft}{\inft} & \lomedup{\inft}{\inft}{\inft} & \lomedup{\inft}{\inft}{\inft}  & \lomedup{\inft}{\inft}{\inft}\\
Car Price & \lomedup{17}{19}{24} & \lomedup{28}{28}{28} & \lomedup{64}{\inft}{\inft} & \lomedup{14}{\textbf{16}}{18} & \lomedup{40}{40}{40}\\
Friedman1 & \lomedup{76}{80}{100} & \lomedup{\inft}{\inft}{\inft} & \lomedup{\inft}{\inft}{\inft} & \lomedup{55}{\textbf{59}}{64} & \lomedup{100}{100}{100}\\
Friedman2 &
\lomedup{13}{\textbf{13}}{13} & \lomedup{19}{19}{19} & \lomedup{23}{27}{33} & \lomedup{14}{16}{17} & \lomedup{40}{40}{40}\\
Friedman3 & \lomedup{\inft}{\inft}{\inft} & \lomedup{\inft}{\inft}{\inft} & \lomedup{\inft}{\inft}{\inft} & \lomedup{73}{\textbf{84}}{\inft} & \lomedup{\inft}{\inft}{\inft}\\
Magic & \lomedup{\inft}{\inft}{\inft} & \lomedup{\inft}{\inft}{\inft} & \lomedup{\inft}{\inft}{\inft} & \lomedup{\inft}{\inft}{\inft} & \lomedup{\inft}{\inft}{\inft}\\
Adult & \lomedup{\inft}{\inft}{\inft} & \lomedup{\inft}{\inft}{\inft} & \lomedup{\inft}{\inft}{\inft} & \lomedup{\inft}{\inft}{\inft} & \lomedup{\inft}{\inft}{\inft}\\
Voice & \lomedup{10}{15}{17} & \lomedup{51}{69}{82} & \lomedup{7}{9}{13} & \lomedup{7}{\textbf{7}}{7} & \lomedup{\inft}{\inft}{\inft}\\
Housing Price & \lomedup{\inft}{\inft}{\inft} & \lomedup{\inft}{\inft}{\inft} & \lomedup{\inft}{\inft}{\inft} & \lomedup{\inft}{\inft}{\inft} & \lomedup{\inft}{\inft}{\inft}\\
\botrule
\textbf{\# Wins} & 1 & 0 & 0 & 6 & 0\\
\end{tabular*}
\end{table}
\begin{table}
\caption{Normalized risk comparison of methods across benchmark datasets, averaged over complexities between 1 to 100 based on \textbf{cross-validation-selected} hyper-parameters. For cases where a method does not achieve a valid solution within this complexity budget, (–) is reported.}\label{table:risk_cv}
\centering
\begin{tabular*}{\textwidth}{@{\extracolsep\fill}lccccc}
\toprule%
& \multicolumn{5}{c}{\rev{Normalized} Train Risks} \\ \cmidrule{2-6}
Dataset & TGB	& NRE & RFObTr & LLTB & EBM \\
\midrule
Iris	& 0.20 \extratiny{$\pm$ 0.09} & 0.45 \extratiny{$\pm$ 0.05} & 0.16 \extratiny{$\pm$ 0.10} & \textbf{0.09} \extratiny{$\pm$ 0.05} & 0.50 \extratiny{$\pm$ 0.07}\\
Liver	& \textbf{0.70} \extratiny{$\pm$ 0.06} & 0.83 \extratiny{$\pm$ 0.00} & \textbf{0.70} \extratiny{$\pm$ 0.04} & 0.78 \extratiny{$\pm$ 0.04} & 0.83 \extratiny{$\pm$ 0.00}\\
Diabetes	& \textbf{0.45} \extratiny{$\pm$ 0.05} & 0.46 \extratiny{$\pm$ 0.02} & 0.46 \extratiny{$\pm$ 0.02} & 0.47 \extratiny{$\pm$ 0.03} & 0.53 \extratiny{$\pm$ 0.00}\\
Breast Cancer	& 0.14 \extratiny{$\pm$ 0.04} & 0.18 \extratiny{$\pm$ 0.03} & \textbf{0.04} \extratiny{$\pm$ 0.01} & 0.08 \extratiny{$\pm$ 0.02} & -\\
Banknote	& 0.22 \extratiny{$\pm$ 0.10} & 0.31 \extratiny{$\pm$ 0.02} & \textbf{0.01} \extratiny{$\pm$ 0.00} & \textbf{0.01} \extratiny{$\pm$ 0.01} & 0.35 \extratiny{$\pm$ 0.06}\\
Red Wine	& 0.64 \extratiny{$\pm$ 0.05} & 0.65 \extratiny{$\pm$ 0.01} & \textbf{0.63} \extratiny{$\pm$ 0.02} & \textbf{0.63} \extratiny{$\pm$ 0.03} & 0.72 \extratiny{$\pm$ 0.00}\\
Car Price	& 0.18 \extratiny{$\pm$ 0.06} & \textbf{0.09} \extratiny{$\pm$ 0.00} & 0.28 \extratiny{$\pm$ 0.02} & 0.14 \extratiny{$\pm$ 0.04} &  0.28 \extratiny{$\pm$ 0.02}\\
Friedman1	& 0.37 \extratiny{$\pm$ 0.10} & 0.42 \extratiny{$\pm$ 0.00} & 0.42 \extratiny{$\pm$ 0.02} & \textbf{0.27} \extratiny{$\pm$ 0.05} & 0.34 \extratiny{$\pm$ 0.00}\\
Friedman2	& 0.12 \extratiny{$\pm$ 0.06} & \textbf{0.02} \extratiny{$\pm$ 0.00} & 0.22 \extratiny{$\pm$ 0.02} & 0.08 \extratiny{$\pm$ 0.04} & 0.26 \extratiny{$\pm$ 0.01}\\
Friedman3	& 0.36 \extratiny{$\pm$ 0.04} & 0.48 \extratiny{$\pm$ 0.05} & 0.36 \extratiny{$\pm$ 0.04} & \textbf{0.32} \extratiny{$\pm$ 0.04} & 0.57 \extratiny{$\pm$ 0.04} \\
Magic	& 0.68 \extratiny{$\pm$ 0.05} & \textbf{0.61} \extratiny{$\pm$ 0.00} & 0.70 \extratiny{$\pm$ 0.03} & 0.62 \extratiny{$\pm$ 0.03} & 0.73 \extratiny{$\pm$ 0.00}\\
Adult	& 0.69 \extratiny{$\pm$ 0.03} & 0.77 \extratiny{$\pm$ 0.05} & 0.66 \extratiny{$\pm$ 0.04} & \textbf{0.62} \extratiny{$\pm$ 0.03} & -\\
Voice	& 0.17 \extratiny{$\pm$ 0.04} & 0.16 \extratiny{$\pm$ 0.00} & \textbf{0.12} \extratiny{$\pm$ 0.01} & \textbf{0.12} \extratiny{$\pm$ 0.02} & -\\
Housing Price	& 0.48 \extratiny{$\pm$ 0.05} & 0.47 \extratiny{$\pm$ 0.02} & 0.49 \extratiny{$\pm$ 0.02} & \textbf{0.39} \extratiny{$\pm$ 0.04} & 0.59 \extratiny{$\pm$ 0.00}\\
%
\midrule
& \multicolumn{5}{c}{\rev{Normalized} Test Risks} \\ \cmidrule{2-6}
Iris	& 0.51 \extratiny{$\pm$ 0.07} & 0.53 \extratiny{$\pm$ 0.02} & 0.42 \extratiny{$\pm$ 0.09} & \textbf{0.40} \extratiny{$\pm$ 0.06} & 0.71 \extratiny{$\pm$ 0.07}\\
liver	& 1.00 \extratiny{$\pm$ 0.01} & 0.96 \extratiny{$\pm$ 0.00} & 0.91 \extratiny{$\pm$ 0.01} & 0.94 \extratiny{$\pm$ 0.01} & \textbf{0.89} \extratiny{$\pm$ 0.00}\\
Diabetes	& 0.72 \extratiny{$\pm$ 0.02} & 0.59 \extratiny{$\pm$ 0.01} & 0.61 \extratiny{$\pm$ 0.02} & 0.61 \extratiny{$\pm$ 0.01} & \textbf{0.57} \extratiny{$\pm$ 0.09}\\
Breast Cancer	& 0.32 \extratiny{$\pm$ 0.02} & \textbf{0.21} \extratiny{$\pm$ 0.02} & \textbf{0.21} \extratiny{$\pm$ 0.02} & 0.22 \extratiny{$\pm$ 0.02} & -\\
Banknote	& 0.26 \extratiny{$\pm$ 0.09} & 0.31 \extratiny{$\pm$ 0.03} & \textbf{0.03} \extratiny{$\pm$ 0.01} & \textbf{0.03} \extratiny{$\pm$ 0.01} & 0.36 \extratiny{$\pm$ 0.06}\\
Red Wine	& 0.75 \extratiny{$\pm$ 0.03} & 0.69 \extratiny{$\pm$ 0.00} & 0.69 \extratiny{$\pm$ 0.01} & \textbf{0.68} \extratiny{$\pm$ 0.02} & 0.73 \extratiny{$\pm$ 0.00} \\
Car Price	& 0.19 \extratiny{$\pm$ 0.05} & \textbf{0.09} \extratiny{$\pm$ 0.00} & 0.29 \extratiny{$\pm$ 0.02} & 0.15 \extratiny{$\pm$ 0.04} & 0.28 \extratiny{$\pm$ 0.02}\\
Friedman1	& 0.40 \extratiny{$\pm$ 0.09} & 0.44 \extratiny{$\pm$ 0.00} & 0.44 \extratiny{$\pm$ 0.02} & \textbf{0.29} \extratiny{$\pm$ 0.05} & 0.34 \extratiny{$\pm$ 0.00}\\
Friedman2	& 0.12 \extratiny{$\pm$ 0.06} & \textbf{0.02} \extratiny{$\pm$ 0.00} & 0.23 \extratiny{$\pm$ 0.02} & 0.08 \extratiny{$\pm$ 0.04} & 0.25 \extratiny{$\pm$ 0.01}\\
Friedman3	& 0.38 \extratiny{$\pm$ 0.04} & 0.50 \extratiny{$\pm$ 0.06} & 0.37 \extratiny{$\pm$ 0.04} & \textbf{0.33} \extratiny{$\pm$ 0.04} & 0.58 \extratiny{$\pm$ 0.04}\\
Magic	& 0.70 \extratiny{$\pm$ 0.04} & \textbf{0.63} \extratiny{$\pm$ 0.00} & 0.72 \extratiny{$\pm$ 0.02} & 0.65 \extratiny{$\pm$ 0.03} & 0.74 \extratiny{$\pm$ 0.00}\\
Adult	& 0.72 \extratiny{$\pm$ 0.02} & 0.77 \extratiny{$\pm$ 0.05} & 0.68 \extratiny{$\pm$ 0.03} & \textbf{0.64} \extratiny{$\pm$ 0.02} & -\\
Voice	& 0.19 \extratiny{$\pm$ 0.04} & 0.16 \extratiny{$\pm$ 0.00} & \textbf{0.15} \extratiny{$\pm$ 0.01} & \textbf{0.15} \extratiny{$\pm$ 0.01} & -\\
Housing Price	& 0.50 \extratiny{$\pm$ 0.05} & 0.47 \extratiny{$\pm$ 0.02} & 0.50 \extratiny{$\pm$ 0.02} & \textbf{0.39} \extratiny{$\pm$ 0.04} & 0.59 \extratiny{$\pm$ 0.00}\\
\botrule
\end{tabular*}
\end{table}
\begin{table}
\caption{Median computation time (in seconds) to achieve 25\% risk reduction (see \Cref{table:complexity_cv}) with dataset size $n'$, dimension $d$, and task type (classification marked (*)); dash indicates that method does not achieve target risk reduction.}\label{table:comp_time}
\centering
\begin{tabular*}{\textwidth}{@{\extracolsep\fill}p{2cm}p{0.07cm}p{0.4cm}ccccc}
\toprule%
Dataset & d	& n$'$ & TGB & NRE & RFObTr & LLTB & EBM \\
\midrule
Iris* & 	4  & 150 & 0.04 & 0.08& 0.04 & 0.08& 7.15  \\
Liver* & 6  & 345 & - &- &- &- & -\\
Diabetes & 10 &  442 & 0.27 & 0.11&  1.71 & 0.25 & 0.22 \\
Breast Cancer* & 30 & 569 & 0.44 & 0.19& 11.09& 0.45 & - \\
Banknote* & 4 &  1372  & 0.16 & 0.52& 0.16 & 0.27 & 0.36 \\
Red Wine & 11 &  1599 & 2.32 & 0.14& 5.25 & 1.71 & 0.26\\
Car Price & 4 &   1770  & 0.23 & 0.12& 0.24 & 0.37 & 0.11 \\
Friedman1 & 10  &  2000 & 1.14 & 0.14& 3.79 & 0.79 & 0.26 \\
Friedman2 & 4 &   3000 & 0.35 & 0.14 & 0.31 & 0.59 & 0.14\\
Friedman3 &  4 & 3000 & 0.39 & 0.15& 0.21 & 0.65 & 0.14 \\
Magic* & 10  & 3000 & 2.62 & -& 8.82 & 2.93 & 7.92 \\ 
Adult* & 13 &  3000 & 1.88 & -&  17.18 & 2.33 & - \\
Voice* & 20 &  3000 & 1.55 & 0.54& 24.76 & 1.38 & - \\
Housing Price & 8 &  3000  & 0.63 & 0.12& 2.24 & 1.66 & 0.35 \\
\botrule
\end{tabular*}
\vspace{-0.8cm}
\end{table}
\section{Conclusion}\label{sec_conclusion}
We introduced LLTBoost, an extension of additive rule ensembles that integrates sparse linear transformations into the gradient boosting framework, enabling oblique decision boundaries with improved expressiveness. Unlike traditional axis-parallel rule learners, LLTBoost builds compact, interpretable rules with explicitly controlled complexity. By formulating rule learning as a sequence of regularized weighted logistic classification problems, the method achieves a comparable time complexity as traditional axis-parallel rule boosting.
Empirical results across 14 benchmark datasets show that LLTBoost outperforms baselines including traditional gradient boosting (TGB), neural rule ensembles (NRE), and RuleFit with oblique trees (RFObTr) in terms of the risk/complexity trade-off. It produces simpler models with equal or better accuracy and relies on a single, interpretable complexity parameter, reducing hyper-parameter tuning efforts.
With this positive evaluation\rev{,} it is important to keep in mind that the complexity measure employed in this work treats axis-parallel and oblique cuts in a very similar way. However, arguably the linear inequalities put a considerably higher cognitive burden on a human interpreter than a simple threshold condition.
To exactly quantify such differences is a subject of interdisciplinary research involving cognition science. 
Still it is possible to address limitations also within the scope of more traditional machine learning research, using sensible rules of thumb. 
For instance, one way to increase the simulatability of linear inequalities is to restrict the value range of the $\bW$-matrix entries to interpretable values such as $w_{i,j} \in \{-5, \dots, 5\}$ as done in so-called ``integer-point scoring systems''~\citep[see, e.g.,][]{ustun2019risk}.
Such a restriction is particularly appealing in conjunction with log transforms, enabling product / ratio representation of simple powers of the input variables.

\backmatter
\rev{\bmhead{Acknowledgments}
This work was supported by the Australian Research Council Laureate Fellowship (FL250100004) and Discovery Project (DP210100045). The authors thank the reviewers for their valuable comments and constructive feedback, which have helped improve the quality of this work.}
\bibliography{cleaned_biblio}
\begin{appendices}
\section{Additional Empirical Results}\label{app1}
To further show the difference between the proposed oblique rule boosting method (LLTBoost) with the other methods, we provide additional results of empirical evaluation.
\subsection{Housing Price Sample Rules}
In the main text, \Cref{table:iris_rules} represents the sample rules for Iris by different methods. Here \Cref{table:housing_rules} represents another set of rule ensembles generated by methods for California housing price dataset. We can also see that the proposed LLTB produces a compact representation of input variables resulting in a less complex ensemble model.
\begin{table}[t]
\centering
\caption{Rules generated by different methods on the California housing price dataset. Each ensemble corresponds to the smallest that achieves at least a 50\% reduction in normalized test risk.}\label{table:housing_rules}
    \begin{tabular}{p{0.05\textwidth}p{0.02\textwidth}p{0.9\textwidth}}
    \toprule
    \multicolumn{3}{l}{\textbf{Traditional Gradient Boosting}} \\
        -0.28 & if & True \\
        +1.28 & if & (\aveoccup $\leq$ 3.42) AND (\medinc$\geq$5.11)\\
        -0.73 & if & \averoom$\geq$4.3 AND \medinc$\leq$2.35\\
        +1.12 & if & (\aveoccup$\leq$2.34) AND (\houseage$\geq$17) AND (\lat$\leq$37.8) AND (\medinc$\geq$3.14)\\ 
        +0.72 & if & (\aveoccup$\leq$2.98) AND (\houseage$\geq$25) AND (\lon$\leq$-118) AND (\medinc$\geq$3.9) \\ 
        -0.49 & if & (\lat$\geq$37.8) AND (\lon$\geq$-122)\\
    \addlinespace[1.5ex]
    \multicolumn{3}{l}{\textbf{Neural Rule Ensembles}} \\
        -0.49 & $\times$ & Min( \ReLU(-1.56\halfspace\medinc -0.31\halfspace \averoom -0.001),  \\
        &&  \ReLU(-0.94\halfspace  \medinc +0.68 \halfspace \averoom -0.02), \\
        && \ReLU(-0.27 \halfspace \medinc -1.44 \halfspace \averoom +0.69) ) \\
        -0.65 & $\times$ & Min( \ReLU(-1.29 \halfspace \medinc +0.94\halfspace  \averoom +0.58), \\
        && \ReLU(-0.36 \halfspace \medinc +0.003\halfspace  \averoom +0.59),\\
        && \ReLU(-0.8 \halfspace \medinc +1.02\halfspace  \averoom +0.33) ) \\
        +0.41 & $\times$ & Min( \ReLU(1.618\halfspace  \medinc +0.34), \\
        && \ReLU(1.24 \halfspace \medinc +1.8), \\
        && \ReLU(0.79 \halfspace \medinc +2.6) )\\
    \addlinespace[1.5ex]
    \multicolumn{3}{l}{\textbf{Rule Fit Oblique Trees}} \\
        +2.33 & if & True \\
        -0.27 & if & (0.16\halfspace \medinc-0.05\halfspace \aveoccup -0.47\halfspace \lat -0.44\halfspace\lon $\leq$ 36.7) \\
        +0.26 & if & (-0.16\halfspace\medinc +0.06\halfspace\aveoccup +0.47\halfspace\halfspace\lat +0.44\halfspace\lon $\leq$ -36.4) \\
        -0.26 &if& (-0.16\halfspace\medinc +0.06\halfspace\aveoccup +0.47\halfspace\lat +0.44\halfspace\lon $>$ -36.4)\\
        -0.44 &if& (0.23\halfspace\medinc +0.006\halfspace\houseage -0.09\halfspace\aveoccup -0.28\halfspace\lat -0.28\halfspace\lon $\leq$ 24.6) \\
        +0.44 &if &(0.23\halfspace\medinc +0.006\halfspace\houseage -0.09\halfspace\aveoccup -0.28\halfspace\lat -0.28\halfspace\lon $>$ 24.6) \\
    \addlinespace[1.5ex]
    \multicolumn{3}{l}{\textbf{Sparse Logistic Linear Transformation Boosting}} \\ 
        -0.59 & if & True\\
        +0.94 & if & (0.52\halfspace\medinc +0.01\halfspace\houseage +0.27\halfspace\avebed -\lat -0.98\halfspace \lon $\geq$ 84.02) \\
        +1.01 & if & (\medinc+0.06\halfspace\houseage $\geq$ 6.7) AND
        (\medinc +0.04\halfspace\houseage-0.13\halfspace\lon $\geq$ 21.3) \\
        \botrule
    \end{tabular}
\end{table}
\subsection{Classification Accuracy Comparison}\label{sec_accuracy}
In this section, we provide additional method comparisons for the classification datasets only in terms of error rate, which corresponds to the empirical $\Risk_{01}$ with respect to the zero–one loss, $\loss_{01}$. 
\rev{Note that, as in previous investigations, normalized risks $\Risk_{01}(c)/\Risk_{01}(0)$ are reported here. This enables a uniform comparison of improvements over the empty model across different datasets (with different class imbalances).}

\rev{Investigating error rates (accuracies) allows to include methods that exclusively work with this metric. In particular, we include here} \textbf{Bayesian Rule Sets (BRS)}~\citep{wang2017bayesian}, which is a popular generate-and-select classifier that learns a sparse rule set by maximizing a Bayesian posterior over a pool of rule conditions generated by association rule mining.
This method of pool construction, which is also utilized in~\citet{lakkaraju2016interpretable}, includes rule conditions based on their coverage (support), which has the disadvantage of requiring very large pool sizes---and thus very large computational budgets---if relatively low coverages are needed to model the target distribution well.
To comply with lower computational budgets, a length constraint can be added.
We use the publicly available implementation\footnote{\url{https://github.com/wangtongada/BOA/tree/master}}, varying rule length in $\{2,3,4\}$ and candidate-pool size in $\{50,100,1000\}$, while other hyper\rev{-}parameters (e.g., rule support and priors) follow the original paper.

\begin{figure}[t]
    \centering
    \includegraphics[width=1\linewidth]{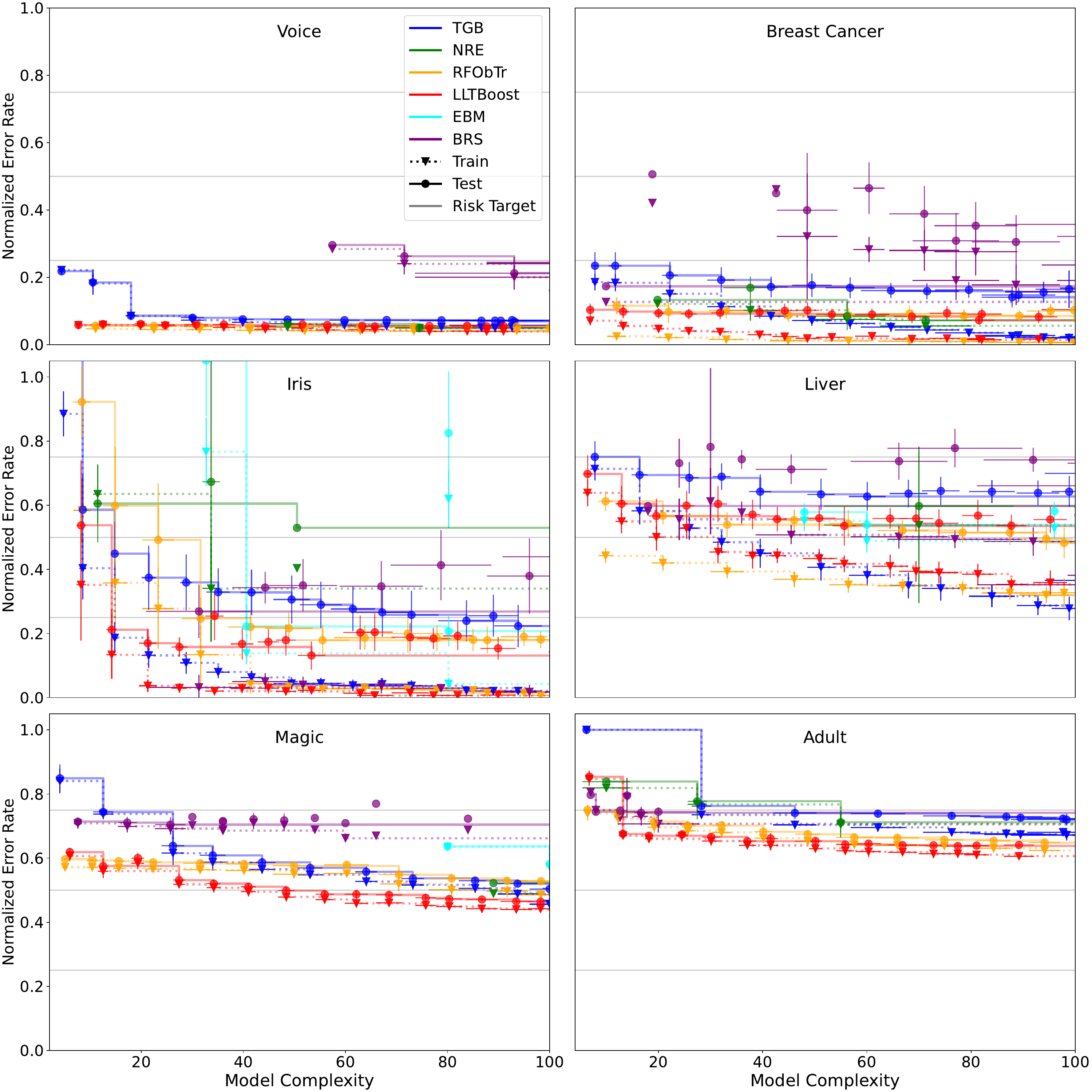}
    \caption{Normalized error rate ($\Risk_{01}(c)/\Risk_{01}(0)$) vs complexity trade-offs by different methods including Bayesian Rule Set (BRS). For each method, ensembles with increasing numbers of rules are extracted, and the mean \rev{error rate} and mean model complexity are reported for each rule count, up to the maximum number of rules corresponding to a model complexity of 100. Error bars represent 95\% CI.}
    \label{fig:complexity_accuracy_trends_append}
\end{figure}

\Cref{fig:complexity_accuracy_trends_append} illustrates the \rev{error rate} versus complexity trade-offs across methods, including BRS, for classification tasks. For each method, ensembles with increasing numbers of rules are extracted, and the mean \rev{error rate} and mean model complexity are reported over 15 repetitions for each rule count, up to the maximum number of rules corresponding to a model complexity of $100$.

\Cref{table:complexity_accuracy} reports, for each method and dataset, the minimum complexity required to achieve error-rate reductions of $25\%$, $50\%$, and $75\%$. As in \Cref{table:complexity_cv}, we report $\infty$ when the target is not achieved or when no solution is observed within the complexity budget of $100$. 
Additionally, \Cref{table:acc_cv} reports normalized classification error rates averaged over \rev{\emph{all}} complexity levels from $1$ to $100$, based on cross-validation-selected hyper-parameters. \rev{Finally, \Cref{table:errorrate_based_on_target_complexity} additionally reports the minimum normalized error rates achievable within complexity of up to $50$ (falling back to the normalized error rate of $1$ achieved by the empty model if no complexity 50 or lower model is produced by the method).}

\rev{In summary, the classification error rate results are mostly aligned with the log loss results. In particular, BRS is only competitive for the two smallest datasets, which can be reasonably attribute to the above-mentioned constraints to the condition candidate pool (required to meet the targeted computational budget).}
\begin{table}
\caption{Model complexity comparison methods across benchmark classification datasets, considering three test error rate reduction (0/1 loss) targets on test normalized error rate based on \textbf{cross-validation-selected} hyper-parameters. For each target, the first ensemble with the smallest complexity that hits the target is selected. Three numbers are
reported for each experiment, representing the median (center), lower (left), and
upper (right) bounds of the 92\% confidence interval.}\label{table:complexity_accuracy}
\centering
\begin{tabular*}{\textwidth}{@{\extracolsep\fill}lcccccc}
\toprule%
& \multicolumn{5}{c}{25\% error rate reduction} \\ \cmidrule{2-7}
Dataset & TGB	& NRE & RFObTr & LLTB & EBM & BRS\\
\midrule
Iris & 
\lomedup{8}{\textbf{8}}{8} & \lomedup{10}{13}{50} & \lomedup{11}{18}{25} & \lomedup{8}{\textbf{8}}{9} & \lomedup{40}{40}{40} & \lomedup{30}{30}{30} \\
Liver & 
\lomedup{7}{9}{18} & \lomedup{\inft}{\inft}{\inft} & \lomedup{8}{9}{11} & \lomedup{5}{\textbf{7}}{8} & \lomedup{36}{36}{36} & \lomedup{36}{42}{48}\\
Breast Cancer & 
\lomedup{7}{7}{7} & \lomedup{36}{50}{66} & \lomedup{7}{13}{13} & \lomedup{6}{\textbf{6}}{7} & \lomedup{\inft}{\inft}{\inft} & \lomedup{57}{63}{69} \\
Banknote & \lomedup{3}{\textbf{3}}{3} & \lomedup{48}{50}{54} & \lomedup{9}{9}{11} & \lomedup{5}{5}{6} & \lomedup{32}{32}{32} & \lomedup{36}{42}{45}\\
Magic & \lomedup{3}{10}{14} & \lomedup{\inft}{\inft}{\inft} & \lomedup{4}{\textbf{5}}{5} & \lomedup{5}{6}{6} & \lomedup{80}{80}{80} & \lomedup{6}{6}{6}\\
Adult & \lomedup{43}{45}{51} & \lomedup{55}{\inft}{\inft} & \lomedup{7}{\textbf{7}}{13} & \lomedup{13}{13}{14} & \lomedup{\inft}{\inft}{\inft} & \lomedup{12}{16}{18} \\
Voice & \lomedup{3}{\textbf{3}}{3} & \lomedup{51}{69}{82} & \lomedup{7}{9}{13} & \lomedup{7}{7}{7} & \lomedup{\inft}{\inft}{\inft} & \lomedup{72}{84}{90} \\
\botrule
\textbf{\# Wins} & 3 & 0 & 2 & 3 & 0 & 0\\
\midrule
& \multicolumn{5}{c}{50\% error rate reduction} \\ \cmidrule{2-7}
Iris & \lomedup{13}{13}{24} & \lomedup{34}{\inft}{\inft} & \lomedup{19}{25}{27} & \lomedup{8}{\textbf{9}}{13} & \lomedup{40}{40}{40} & \lomedup{30}{30}{36}\\
Liver & \lomedup{\inft}{\inft}{\inft} & \lomedup{\inft}{\inft}{\inft} & \lomedup{54}{63}{89} & \lomedup{45}{\textbf{58}}{\inft} & \lomedup{\inft}{\inft}{\inft} & \lomedup{\inft}{\inft}{\inft}\\
Breast Cancer & 
\lomedup{7}{7}{7} & \lomedup{36}{50}{66} & \lomedup{7}{13}{13} & \lomedup{6}{\textbf{6}}{7} & \lomedup{\inft}{\inft}{\inft} & \lomedup{57}{69}{72}\\
Banknote & \lomedup{3}{\textbf{3}}{3} & \lomedup{48}{50}{54} & \lomedup{9}{9}{11} & \lomedup{5}{5}{6} & \lomedup{32}{32}{32} & \lomedup{42}{42}{48}\\
Magic & \lomedup{\inft}{\inft}{\inft} & \lomedup{\inft}{\inft}{\inft} & \lomedup{\inft}{\inft}{\inft} & \lomedup{40}{\textbf{43}}{47} & \lomedup{\inft}{\inft}{\inft} & \lomedup{\inft}{\inft}{\inft}\\
Adult & \lomedup{\inft}{\inft}{\inft} & \lomedup{\inft}{\inft}{\inft} & \lomedup{\inft}{\inft}{\inft} & \lomedup{\inft}{\inft}{\inft} & \lomedup{\inft}{\inft}{\inft} & \lomedup{\inft}{\inft}{\inft}\\
Voice & \lomedup{3}{\textbf{3}}{3} & \lomedup{51}{69}{82} & \lomedup{7}{9}{13} & \lomedup{7}{7}{7} & \lomedup{\inft}{\inft}{\inft} & \lomedup{72}{84}{90} \\
\botrule
\textbf{\# Wins} & 2 & 0 & 0 & 4 & 0 & 0\\
\botrule
& \multicolumn{5}{c}{75\% error rate reduction} \\ \cmidrule{2-7}
Iris & \lomedup{16}{44}{89} & \lomedup{\inft}{\inft}{\inft} & \lomedup{25}{28}{36} & \lomedup{9}{\textbf{12}}{16} & \lomedup{40}{40}{80} & \lomedup{36}{42}{\inft}\\
Liver & \lomedup{\inft}{\inft}{\inft} & \lomedup{\inft}{\inft}{\inft} & \lomedup{\inft}{\inft}{\inft} & \lomedup{\inft}{\inft}{\inft} & \lomedup{\inft}{\inft}{\inft} & \lomedup{\inft}{\inft}{\inft}\\
Breast Cancer & 
\lomedup{7}{7}{32} & \lomedup{36}{50}{66} & \lomedup{7}{13}{13} & \lomedup{6}{\textbf{6}}{7} & \lomedup{\inft}{\inft}{\inft} & \lomedup{90}{\inft}{\inft} \\
Banknote & \lomedup{24}{26}{26} & \lomedup{50}{53}{54} & \lomedup{9}{9}{11} & \lomedup{5}{\textbf{5}}{6} & \lomedup{40}{40}{40} & \lomedup{48}{51}{54} \\
Magic & \lomedup{\inft}{\inft}{\inft} & \lomedup{\inft}{\inft}{\inft} & \lomedup{\inft}{\inft}{\inft} & \lomedup{\inft}{\inft}{\inft} & \lomedup{\inft}{\inft}{\inft} & \lomedup{\inft}{\inft}{\inft}\\
Adult & \lomedup{\inft}{\inft}{\inft} & \lomedup{\inft}{\inft}{\inft} & \lomedup{\inft}{\inft}{\inft} & \lomedup{\inft}{\inft}{\inft} & \lomedup{\inft}{\inft}{\inft} & \lomedup{\inft}{\inft}{\inft}\\
Voice & \lomedup{3}{\textbf{3}}{3} & \lomedup{51}{69}{82} & \lomedup{7}{9}{13} & \lomedup{7}{7}{7} & \lomedup{\inft}{\inft}{\inft} & \lomedup{90}{\inft}{\inft} \\
\botrule
\textbf{\# Wins} & 1 & 0 & 0 & 3 & 0 & 0\\
\end{tabular*}
\end{table}
\begin{table}[t]
\caption{\rev{Normalized Error rate} comparison of methods across benchmark classification datasets, averaged over complexities between 1 to 100 based on \textbf{cross-validation-selected} hyper-parameters. For cases where a method does not achieve a valid solution within this complexity budget, (–) is reported.}\label{table:acc_cv}
\centering
\begin{tabular*}{\textwidth}{@{\extracolsep\fill}lcccccc}
\toprule%
& \multicolumn{6}{c}{Normalized Train Error Rates} \\ \cmidrule{2-7}
Dataset & TGB	& NRE & RFObTr & LLTB & EBM & BRS \\
\midrule
Iris	& 0.14 \extratiny{$\pm$ 0.11} & 0.46 \extratiny{$\pm$ 0.06} & 0.11 \extratiny{$\pm$ 0.08} & 0.05 \extratiny{$\pm$ 0.04} & 0.47 \extratiny{$\pm$ 0.08} & \textbf{0.03} \extratiny{$\pm$ 0.01}  \\
Liver	& 0.41 \extratiny{$\pm$ 0.06} & 0.42 \extratiny{$\pm$ 0.04} & 0.36 \extratiny{$\pm$ 0.02} & 0.45 \extratiny{$\pm$ 0.04} & 0.52 \extratiny{$\pm$ 0.00} & \textbf{0.34} \extratiny{$\pm$ 0.09}  \\
Breast Cancer	& 0.07 \extratiny{$\pm$ 0.03} & 0.09 \extratiny{$\pm$ 0.01} & \textbf{0.01} \extratiny{$\pm$ 0.00} & 0.03 \extratiny{$\pm$ 0.01} & - & 0.24 \extratiny{$\pm$ 0.05} \\
Banknote	& 0.13 \extratiny{$\pm$ 0.06} & 0.22 \extratiny{$\pm$ 0.02} & \textbf{0.00} \extratiny{$\pm$ 0.00} & \textbf{0.00} \extratiny{$\pm$ 0.00} & 0.21 \extratiny{$\pm$ 0.03} & 0.37 \extratiny{$\pm$ 0.06}  \\
Magic	& 0.57 \extratiny{$\pm$ 0.06} & \textbf{0.49} \extratiny{$\pm$ 0.00} & 0.54 \extratiny{$\pm$ 0.02} & \textbf{0.49} \extratiny{$\pm$ 0.03} & 0.60 \extratiny{$\pm$ 0.00} & 0.69 \extratiny{$\pm$ 0.01} \\
Adult	& 0.72 \extratiny{$\pm$ 0.06} & 0.76 \extratiny{$\pm$ 0.03} & 0.67 \extratiny{$\pm$ 0.02} & \textbf{0.65} \extratiny{$\pm$ 0.04} & - & 0.75 \extratiny{$\pm$ 0.02}  \\
Voice	& 0.08 \extratiny{$\pm$ 0.03} & 0.05 \extratiny{$\pm$ 0.00} & \textbf{0.04} \extratiny{$\pm$ 0.00} & 0.05 \extratiny{$\pm$ 0.00} & - & 0.19 \extratiny{$\pm$ 0.02} \\
\botrule
\textbf{\# Wins} &0 & 1&  3&  6 &\\
\midrule
& \multicolumn{6}{c}{Normalized Test Error Rates} \\ \cmidrule{2-7}
Iris	& 0.38 \extratiny{$\pm$ 0.12} & 0.60 \extratiny{$\pm$ 0.03} & 0.29 \extratiny{$\pm$ 0.11} & \textbf{0.21} \extratiny{$\pm$ 0.05} & 0.66 \extratiny{$\pm$ 0.08} & 0.35 \extratiny{$\pm$ 0.04} \\
Liver	& 0.67 \extratiny{$\pm$ 0.02} & 0.65 \extratiny{$\pm$ 0.04} & \textbf{0.53} \extratiny{$\pm$ 0.02} & 0.57 \extratiny{$\pm$ 0.02} & 0.56 \extratiny{$\pm$ 0.00} & 0.69 \extratiny{$\pm$ 0.03} \\
Breast Cancer	& 0.18 \extratiny{$\pm$ 0.01} & 0.11 \extratiny{$\pm$ 0.02} & 0.10 \extratiny{$\pm$ 0.00} & \textbf{0.09} \extratiny{$\pm$ 0.00} & - & 0.35 \extratiny{$\pm$ 0.05}\\
Banknote	& 0.15 \extratiny{$\pm$ 0.06} & 0.23 \extratiny{$\pm$ 0.02} & \textbf{0.01} \extratiny{$\pm$ 0.00} & \textbf{0.01} \extratiny{$\pm$ 0.00} & 0.22 \extratiny{$\pm$ 0.03} & 0.41 \extratiny{$\pm$ 0.06} \\
Magic	& 0.60 \extratiny{$\pm$ 0.06} & 0.52 \extratiny{$\pm$ 0.00} & 0.57 \extratiny{$\pm$ 0.01} & \textbf{0.51} \extratiny{$\pm$ 0.03} & 0.61 \extratiny{$\pm$ 0.00} & 0.72 \extratiny{$\pm$ 0.01} \\
Adult	& 0.76 \extratiny{$\pm$ 0.05} & 0.78 \extratiny{$\pm$ 0.03} & 0.69 \extratiny{$\pm$ 0.02} & \textbf{0.67} \extratiny{$\pm$ 0.03} & - & 0.76 \extratiny{$\pm$ 0.02} \\
Voice	& 0.09 \extratiny{$\pm$ 0.02} & \textbf{0.05} \extratiny{$\pm$ 0.00} & \textbf{0.05} \extratiny{$\pm$ 0.00} & 0.06 \extratiny{$\pm$ 0.00} & - & 0.21 \extratiny{$\pm$ 0.02}  \\
\botrule
\end{tabular*}
\end{table}
\begin{table}[t]
\caption{\rev{Normalized error rate comparison of methods across benchmark datasets for a target model complexity of $50$, based on \textbf{cross-validation-selected} hyper-parameters. Three numbers are
reported for each experiment, representing the median (center), lower (left), and
upper (right) bounds of the 92\% confidence interval.}}\label{table:errorrate_based_on_target_complexity}
\centering
\begin{tabular*}{1.05\textwidth}{@{\extracolsep\fill}lcccccc}
\toprule%
& \multicolumn{6}{c}{Normalized Train Error Rate} \\ \cmidrule{2-7}
Dataset & TGB	& NRE & RFObTr & LLTB & EBM & BRS\\
\midrule
Iris & \lomedup{0.02}{0.04}{0.06} & \lomedup{0.52}{0.53}{0.67} & \lomedup{0.02}{\textbf{0.02}}{0.05} & \lomedup{0}{\textbf{0.02}}{0.05} & \lomedup{0.09}{0.13}{0.15} & \lomedup{0.02}{0.04}{0.05}\\
Liver & \lomedup{0.37}{0.46}{0.49} & \lomedup{1}{1}{1} & \lomedup{0.34}{\textbf{0.36}}{0.41} & \lomedup{0.4}{0.45}{0.47} & \lomedup{0.53}{0.54}{0.57} & \lomedup{0.5}{0.56}{0.57}\\
Breast Cancer & \lomedup{0.07}{0.08}{0.1} & \lomedup{0.11}{0.15}{1} & \lomedup{0.01}{\textbf{0.01}}{0.01} & \lomedup{0.02}{0.02}{0.03} & \lomedup{1}{1}{1} & \lomedup{1}{1}{1}\\
Banknote & \lomedup{0.08}{0.09}{0.12} & \lomedup{0.21}{0.31}{1} & \lomedup{0}{\textbf{0}}{0} & \lomedup{0}{\textbf{0}}{0} & \lomedup{0.15}{0.16}{0.18} & \lomedup{0.17}{0.4}{0.46}\\
Magic & \lomedup{0.54}{0.55}{0.57} & \lomedup{1}{1}{1} & \lomedup{0.54}{0.57}{0.57} & \lomedup{0.47}{\textbf{0.47}}{0.49} & \lomedup{1}{1}{1} & \lomedup{0.68}{0.69}{0.71}\\
Adult & \lomedup{0.69}{0.7}{0.72} & \lomedup{0.75}{0.76}{0.82} & \lomedup{0.64}{0.66}{0.67} & \lomedup{0.62}{\textbf{0.62}}{0.63} & \lomedup{1}{1}{1} & \lomedup{0.72}{0.73}{0.74}\\
Voice & \lomedup{0.06}{0.06}{0.07} & \lomedup{1}{1}{1} & \lomedup{0.04}{\textbf{0.04}}{0.05} & \lomedup{0.04}{0.05}{0.05} & \lomedup{1}{1}{1} & \lomedup{1}{1}{1}\\
\midrule
& \multicolumn{6}{c}{Normalized Test Risk} \\ \cmidrule{2-7}
Dataset & TGB	& NRE & RFObTr & LLTB & EBM & BRS\\
\midrule
Iris & \lomedup{0.19}{0.33}{0.4} & \lomedup{0.47}{0.56}{0.61} & \lomedup{0.15}{0.2}{0.24} & \lomedup{0.11}{\textbf{0.14}}{0.2} & \lomedup{0.13}{0.19}{0.29} & \lomedup{0.24}{0.29}{0.39}\\
Liver & \lomedup{0.61}{0.65}{0.72} & \lomedup{1}{1}{1} & \lomedup{0.51}{\textbf{0.52}}{0.57} & \lomedup{0.5}{0.57}{0.6} & \lomedup{0.55}{0.6}{0.61} & \lomedup{0.68}{0.72}{0.74}\\
Breast Cancer & \lomedup{0.14}{0.17}{0.21} & \lomedup{0.15}{0.22}{1} & \lomedup{0.07}{0.11}{0.11} & \lomedup{0.08}{\textbf{0.09}}{0.1} & \lomedup{1}{1}{1} & \lomedup{1}{1}{1}\\
Banknote & \lomedup{0.08}{0.1}{0.13} & \lomedup{0.21}{0.3}{1} & \lomedup{0}{0.01}{0.01} & \lomedup{0}{\textbf{0}}{0.01} & \lomedup{0.16}{0.17}{0.18} & \lomedup{0.25}{0.44}{0.5}\\
Magic & \lomedup{0.57}{0.58}{0.59} & \lomedup{1}{1}{1} & \lomedup{0.58}{0.59}{0.59} & \lomedup{0.48}{\textbf{0.5}}{0.51} & \lomedup{1}{1}{1} & \lomedup{0.7}{0.72}{0.73}\\
Adult & \lomedup{0.73}{0.74}{0.75} & \lomedup{0.75}{0.76}{0.84} & \lomedup{0.65}{0.68}{0.7} & \lomedup{0.65}{\textbf{0.65}}{0.65} & \lomedup{1}{1}{1} & \lomedup{0.74}{0.74}{0.76}\\
Voice & \lomedup{0.07}{0.07}{0.08} & \lomedup{1}{1}{1} & \lomedup{0.05}{\textbf{0.05}}{0.06} & \lomedup{0.05}{0.06}{0.06} & \lomedup{1}{1}{1} & \lomedup{1}{1}{1}\\
\botrule
\end{tabular*}
\end{table}
\subsection{Additional Risk/Complexity Curves \rev{and Risk Table}}
\label{sec_additional_curves}
\begin{figure}[t]
    \centering
    \includegraphics[width=1\linewidth]{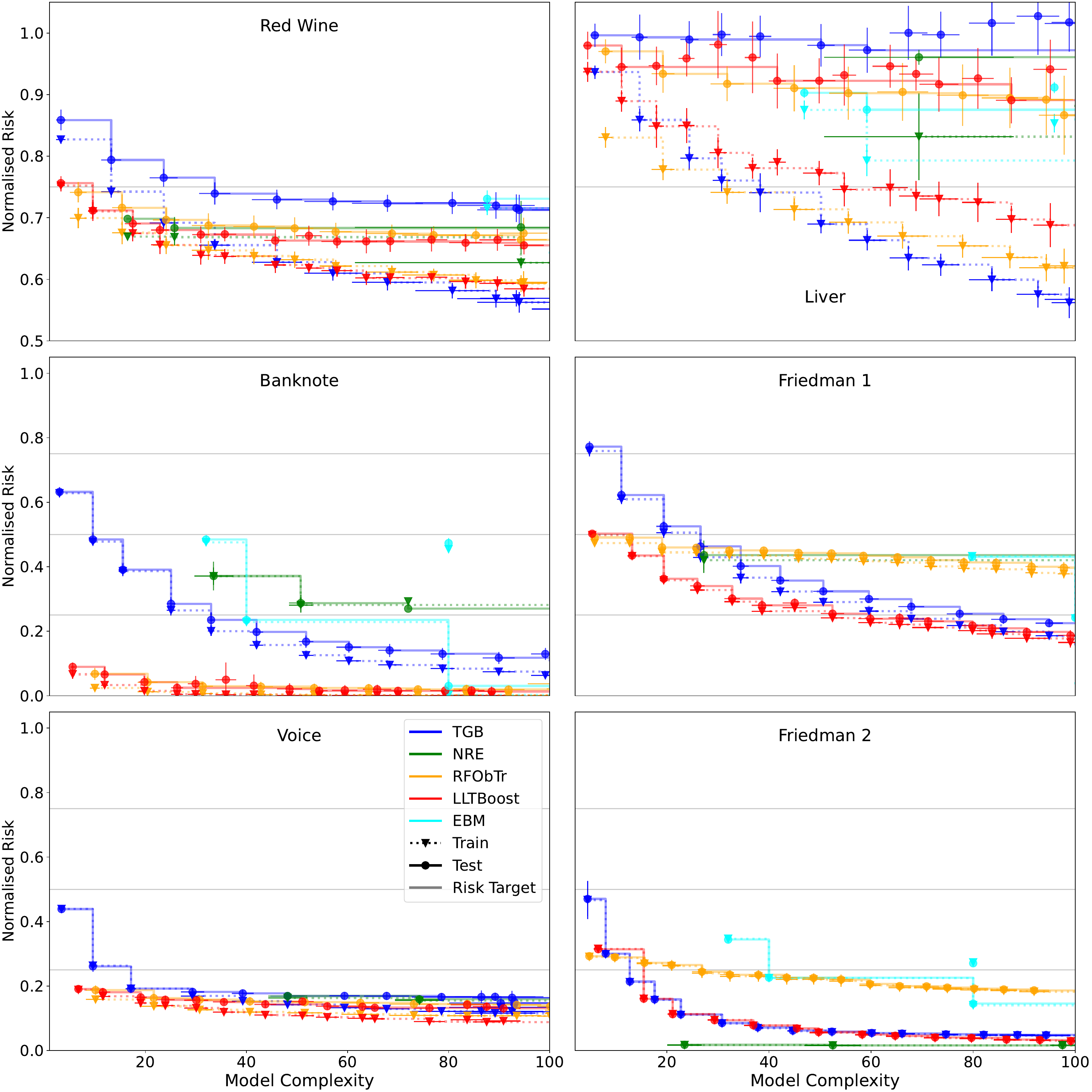}
    \caption{Risk vs complexity trade-offs by different methods for \texttt{red wine}, \texttt{liver}, \texttt{banknote}, \texttt{voice}, \texttt{Friedman1}, \texttt{Friedman2} datasets based on \textbf{cross-validation-selected} hyper-parameters, continuing \Cref{fig:complexity_risk_cv_trends}. For each method, ensembles with increasing numbers of rules are extracted, and the mean normalized risk and mean model complexity are reported over 15 repetitions for each rule count, up to the maximum number of rules corresponding to a model complexity of 100. Error bars represent 95\% CI.}
    \label{fig:complexity_risk_cv_trends_append}
\end{figure}

Due to space limitations, \Cref{fig:complexity_risk_cv_trends} in the main text presents risk/complexity trade-offs for only eight datasets: \texttt{adult}, \texttt{magic}, \texttt{housing price}, \texttt{friedman3}, \texttt{used car price}, \texttt{iris}, \texttt{diabetes}, and \texttt{breast cancer}. Here, \Cref{fig:complexity_risk_cv_trends_append} provides the corresponding curves for six additional datasets, based on cross-validation–selected hyper-parameters.
The same overall pattern is observed: LLTB achieves a favorable risk/complexity trade-off. In some datasets, such as \texttt{red wine}, \texttt{banknote}, and \texttt{voice}, RFObTr is also highly competitive, highlighting the benefit of linear representation learning in certain settings.

\rev{In addition, \Cref{table:risk_based_on_target_complexity} reports the minimum risk achievable within complexity of up to $50$ (falling back to the normalized error rate of $1$ achieved by the empty model if no complexity 50 or lower model is produced by the method). As can be seen, this also confirms the results in \Cref{table:risk_cv}, where LLTBoost achieved the lowest risks on the majority of datasets compared to the other methods.}
\begin{table}[t]
\caption{\rev{Normalized risk comparison of methods across benchmark datasets for a target model complexity of $50$, based on \textbf{cross-validation-selected} hyper-parameters. Three numbers are
reported for each experiment, representing the median (center), lower (left), and
upper (right) bounds of the 92\% confidence interval.}}\label{table:risk_based_on_target_complexity}
\centering
\begin{tabular*}{\textwidth}{@{\extracolsep\fill}lccccc}
\toprule%
& \multicolumn{5}{c}{Normalized Train Risk} \\ \cmidrule{2-6}
Dataset & TGB	& NRE & RFObTr & LLTB & EBM\\
\midrule
Iris & \lomedup{0.09}{0.1}{0.12} & \lomedup{0.55}{0.57}{0.59} & \lomedup{0.06}{0.08}{0.1} & \lomedup{0.04}{\textbf{0.07}}{0.07} & \lomedup{0.18}{0.2}{0.25}\\
Liver & \lomedup{0.7}{\textbf{0.71}}{0.73} & \lomedup{1}{1}{1} & \lomedup{0.69}{0.72}{0.75} & \lomedup{0.76}{0.77}{0.8} & \lomedup{0.85}{0.88}{0.89}\\
Diabetes & \lomedup{0.46}{0.48}{0.5} & \lomedup{0.48}{0.5}{0.51} & \lomedup{0.43}{0.45}{0.47} & \lomedup{0.44}{\textbf{0.44}}{0.47} & \lomedup{1}{1}{1}\\
Breast Cancer & \lomedup{0.12}{0.13}{0.14} & \lomedup{0.18}{0.27}{1} & \lomedup{0.03}{\textbf{0.03}}{0.04} & \lomedup{0.05}{0.07}{0.08} & \lomedup{1}{1}{1}\\
Banknote & \lomedup{0.13}{0.15}{0.17} & \lomedup{0.28}{0.39}{1} & \lomedup{0}{\textbf{0}}{0} & \lomedup{0}{\textbf{0}}{0} & \lomedup{0.22}{0.23}{0.24}\\
Red Wine & \lomedup{0.62}{0.64}{0.65} & \lomedup{0.66}{0.67}{0.68} & \lomedup{0.63}{0.64}{0.65} & \lomedup{0.6}{\textbf{0.62}}{0.63} & \lomedup{1}{1}{1}\\
Car Price & \lomedup{0.13}{0.14}{0.14} & \lomedup{0.09}{\textbf{0.1}}{0.1} & \lomedup{0.26}{0.27}{0.28} & \lomedup{0.11}{0.12}{0.13} & \lomedup{0.22}{0.23}{0.23}\\
Friedman 1 & \lomedup{0.29}{0.31}{0.32} & \lomedup{0.33}{0.46}{0.47} & \lomedup{0.41}{0.42}{0.43} & \lomedup{0.25}{\textbf{0.26}}{0.27} & \lomedup{1}{1}{1}\\
Friedman 2 & \lomedup{0.06}{0.06}{0.06} & \lomedup{0.02}{\textbf{0.02}}{0.02} & \lomedup{0.2}{0.21}{0.22} & \lomedup{0.06}{0.06}{0.07} & \lomedup{0.22}{0.22}{0.23}\\
Friedman 3 & \lomedup{0.33}{0.33}{0.34} & \lomedup{0.44}{0.45}{0.46} & \lomedup{0.35}{0.37}{0.42} & \lomedup{0.31}{\textbf{0.32}}{0.33} & \lomedup{0.51}{0.52}{0.52}\\
Magic & \lomedup{0.65}{0.67}{0.68} & \lomedup{1}{1}{1} & \lomedup{0.7}{0.7}{0.72} & \lomedup{0.6}{\textbf{0.61}}{0.61} & \lomedup{1}{1}{1}\\
Adult & \lomedup{0.7}{0.7}{0.71} & \lomedup{0.75}{0.76}{0.77} & \lomedup{0.63}{0.63}{0.66} & \lomedup{0.6}{\textbf{0.61}}{0.61} & \lomedup{1}{1}{1}\\
Voice & \lomedup{0.14}{0.15}{0.15} & \lomedup{1}{1}{1} & \lomedup{0.11}{0.12}{0.13} & \lomedup{0.11}{\textbf{0.11}}{0.12} & \lomedup{1}{1}{1}\\
Housing Price & \lomedup{0.44}{0.45}{0.46} & \lomedup{0.51}{0.51}{0.52} & \lomedup{0.46}{0.48}{0.5} & \lomedup{0.35}{\textbf{0.36}}{0.44} & \lomedup{1}{1}{1}\\
\midrule
& \multicolumn{5}{c}{Normalized Test Risk} \\ \cmidrule{2-6}
Dataset & TGB	& NRE & RFObTr & LLTB & EBM\\
\midrule
Iris & \lomedup{0.24}{\textbf{0.32}}{0.64} & \lomedup{0.52}{0.56}{0.6} & \lomedup{0.29}{0.33}{0.4} & \lomedup{0.22}{\textbf{0.32}}{0.38} & \lomedup{0.29}{0.42}{0.48}\\
Liver & \lomedup{0.94}{1.01}{1.03} & \lomedup{1}{1}{1} & \lomedup{0.89}{0.91}{0.94} & \lomedup{0.92}{0.92}{0.97} & \lomedup{0.9}{\textbf{0.9}}{0.91}\\
Diabetes & \lomedup{0.67}{0.7}{0.73} & \lomedup{0.53}{\textbf{0.56}}{0.62} & \lomedup{0.57}{0.6}{0.65} & \lomedup{0.59}{0.61}{0.65} & \lomedup{1}{1}{1}\\
Breast Cancer & \lomedup{0.26}{0.29}{0.32} & \lomedup{0.26}{0.31}{1} & \lomedup{0.15}{0.21}{0.24} & \lomedup{0.15}{\textbf{0.18}}{0.25} & \lomedup{1}{1}{1}\\
Banknote & \lomedup{0.16}{0.17}{0.19} & \lomedup{0.29}{0.4}{1} & \lomedup{0.01}{0.02}{0.03} & \lomedup{0}{\textbf{0.01}}{0.01} & \lomedup{0.22}{0.23}{0.25}\\
Red Wine & \lomedup{0.71}{0.73}{0.76} & \lomedup{0.65}{0.69}{0.7} & \lomedup{0.66}{0.68}{0.69} & \lomedup{0.66}{\textbf{0.66}}{0.68} & \lomedup{1}{1}{1}\\
Car Price & \lomedup{0.14}{0.15}{0.15} & \lomedup{0.09}{\textbf{0.1}}{0.1} & \lomedup{0.26}{0.28}{0.29} & \lomedup{0.12}{0.14}{0.15} & \lomedup{0.23}{0.23}{0.24}\\
Friedman 1 & \lomedup{0.33}{0.34}{0.35} & \lomedup{0.36}{0.48}{0.5} & \lomedup{0.43}{0.45}{0.46} & \lomedup{0.27}{\textbf{0.28}}{0.29} & \lomedup{1}{1}{1}\\
Friedman 2 & \lomedup{0.06}{0.06}{0.06} & \lomedup{0.02}{\textbf{0.02}}{0.02} & \lomedup{0.21}{0.22}{0.23} & \lomedup{0.06}{0.07}{0.07} & \lomedup{0.22}{0.23}{0.23}\\
Friedman 3 & \lomedup{0.33}{0.34}{0.35} & \lomedup{0.45}{0.46}{0.46} & \lomedup{0.36}{0.38}{0.41} & \lomedup{0.31}{\textbf{0.32}}{0.34} & \lomedup{0.52}{0.52}{0.52}\\
Magic & \lomedup{0.68}{0.68}{0.7} & \lomedup{1}{1}{1} & \lomedup{0.72}{0.73}{0.74} & \lomedup{0.63}{\textbf{0.63}}{0.64} & \lomedup{1}{1}{1}\\
Adult & \lomedup{0.72}{0.73}{0.73} & \lomedup{0.75}{0.76}{0.81} & \lomedup{0.65}{0.65}{0.66} & \lomedup{0.62}{\textbf{0.62}}{0.63} & \lomedup{1}{1}{1}\\
Voice & \lomedup{0.16}{0.17}{0.18} & \lomedup{1}{1}{1} & \lomedup{0.13}{0.15}{0.17} & \lomedup{0.13}{\textbf{0.14}}{0.16} & \lomedup{1}{1}{1}\\
Housing Price & \lomedup{0.47}{0.47}{0.49} & \lomedup{0.52}{0.52}{0.52} & \lomedup{0.48}{0.49}{0.49} & \lomedup{0.36}{\textbf{0.37}}{0.42} & \lomedup{1}{1}{1}\\
\botrule
\end{tabular*}
\end{table}
\subsection{Risk/Complexity Evaluation with Oracle Selection}\label{sec_oracle_results}
In the main text and \Cref{sec_additional_curves}, methods are compared with respect to hyper-parameter values learned from the training data.
While this reflects the performances of these methods when used in practice, suboptimal hyper-parameter choices can obfuscate the underlying trends in the risk/complexity curves---in particular render them not monotonically decreasing due to overfitting.
This is a general concern for all methods, which is, however, not the main focus of our study. Therefore, it is interesting to also investigate idealized ``oracle'' performances using the best hyper-parameter values in hindsight.

\begin{figure}[t]
    \centering
    \includegraphics[width=1\textwidth]{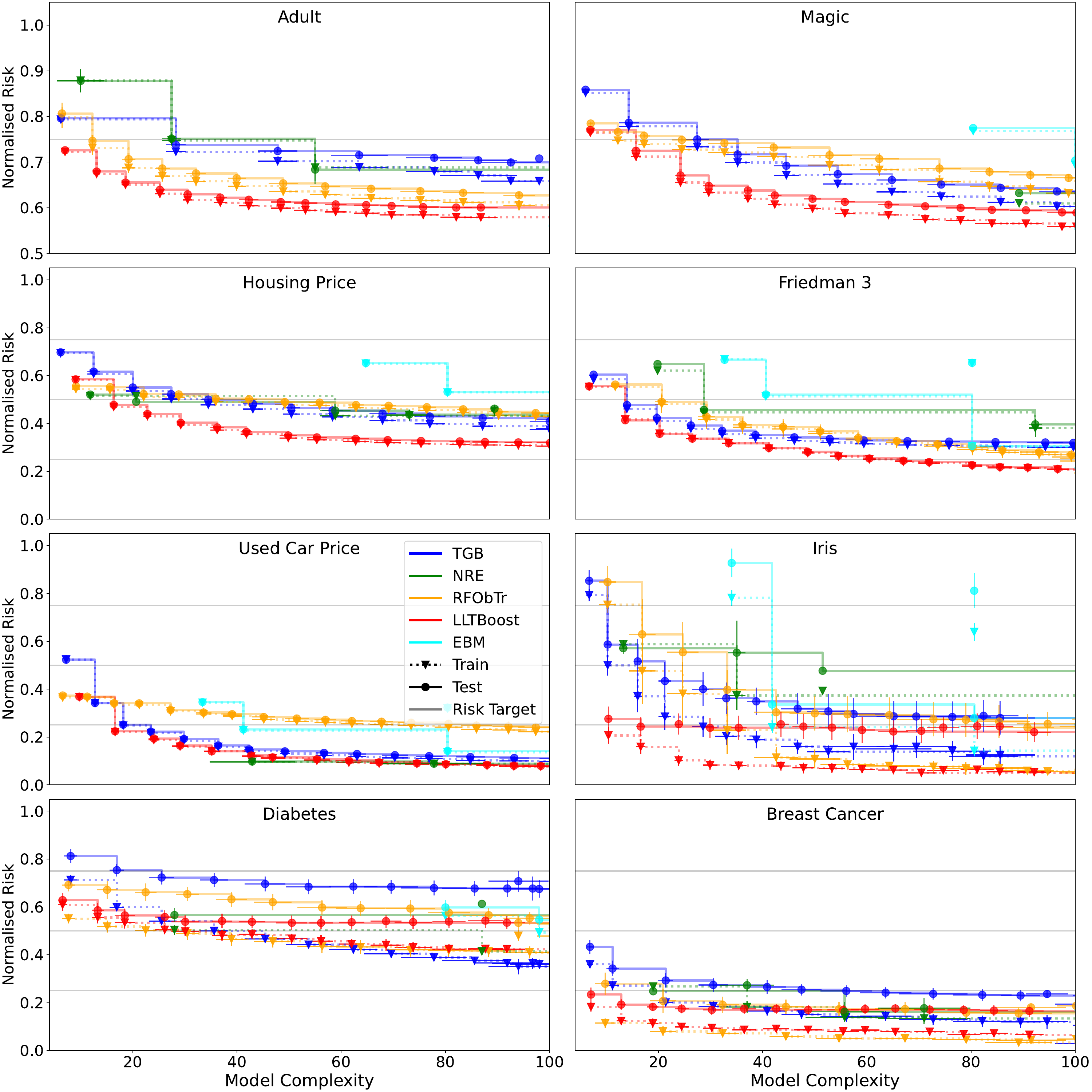}
    \vspace{-0.2cm}
    \caption{Risk vs complexity trade-offs by different methods for \texttt{adult}, \texttt{magic}, \texttt{housing price}, \texttt{Friedman3}, \texttt{used car price}, \texttt{iris}, \texttt{diabetes}, \texttt{breast cancer} datasets based on \textbf{oracle-selected} hyper-parameters. For each method, ensembles with increasing numbers of rules are extracted, and the mean normalized risk and mean model complexity are reported over 15 repetitions for each rule count, up to the maximum number of rules corresponding to a model complexity of 100. Error bars represent 95\% CI.}
    \label{fig:complexity_risk_oracle_trends_1}
\end{figure}

\begin{figure}[t]
    \centering
    \includegraphics[width=1\textwidth]{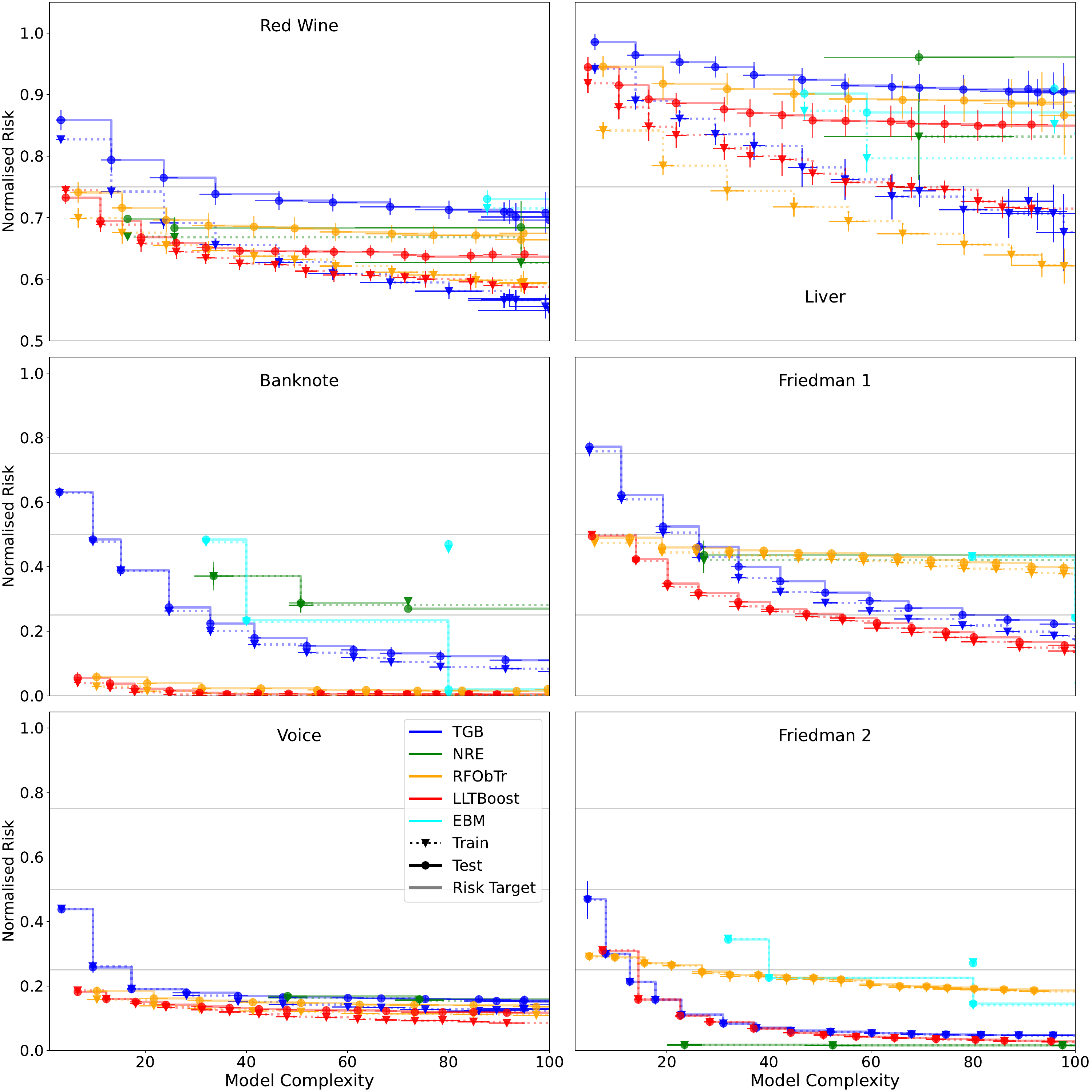}
    \vspace{-0.2cm}
    \caption{Risk vs complexity trade-offs by different methods for \texttt{red wine}, \texttt{liver}, \texttt{banknote}, \texttt{voice}, \texttt{Friedman1}, \texttt{Friedman2} datasets based on \textbf{oracle-selected} hyper-parameters, continuing \Cref{fig:complexity_risk_oracle_trends_1}. For each method, ensembles with increasing numbers of rules are extracted, and the mean normalized risk and mean model complexity are reported over 15 repetitions for each rule count, up to the maximum number of rules corresponding to a model complexity of 100. Error bars represent 95\% CI.}
    \label{fig:complexity_risk_oracle_trends_2}
\end{figure}

\Cref{table:complexity_oracle,table:risk_oracle} and \Cref{fig:complexity_risk_oracle_trends_1,fig:complexity_risk_oracle_trends_2} present the corresponding comparisons of complexity, risk, and error rate obtained under oracle selection of the $\lambda_{l2}$ regularization parameter where for each repetition the test-risk minimizing $\lambda_{l2}$-values for TGB, RFObTr, LLTB, and EBM are chosen.
The figures show overall similar trends as the results with respect to learned hyper-parameter values, though as intended they are more pronounced and, in particular, the risk curves are monotonically decreasing in the model complexity.
Specifically, the difference is strongest for the smallest datasets where learning good hyper-parameter values is particularly challenging.
\begin{table}[t]
\caption{Model complexity comparison of methods across benchmark datasets, considering three test risk reduction targets on normalized test risk based on \textbf{oracle-selected} hyper-parameters. For each target, the first ensemble with the smallest complexity that hits the target is selected. Three numbers are
reported for each experiment, representing the median (center), lower (left), and
upper (right) bounds of the 92\% confidence interval over 15 repetitions.}\label{table:complexity_oracle}
\centering
\begin{tabular*}{\textwidth}{@{\extracolsep\fill}lccccc}
\toprule%
& \multicolumn{5}{c}{25\% risk reduction} \\ \cmidrule{2-6}
Dataset & TGB	& NRE & RFObTr & LLTB & EBM\\
\midrule
Iris & 
\lomedup{6}{8}{8} & \lomedup{10}{10}{13} & \lomedup{15}{21}{26} & \lomedup{6}{\textbf{7}}{7} & \lomedup{40}{40}{40}\\
Liver & \lomedup{\inft}{\inft}{\inft} & \lomedup{\inft}{\inft}{\inft} & \lomedup{\inft}{\inft}{\inft} & \lomedup{\inft}{\inft}{\inft} & \lomedup{\inft}{\inft}{\inft}\\
Diabetes & 
\lomedup{14}{16}{31} & \lomedup{28}{28}{28} & \lomedup{6}{7}{9} & \lomedup{4}{\textbf{4}}{4} & \lomedup{80}{80}{80}\\
Breast Cancer & 
\lomedup{7}{7}{7} & \lomedup{36}{50}{66} & \lomedup{7}{7}{13} & \lomedup{5}{\textbf{5}}{5} & \lomedup{\inft}{\inft}{\inft}\\
Banknote & \lomedup{3}{\textbf{3}}{3} & \lomedup{48}{50}{54} & \lomedup{9}{9}{11} & \lomedup{5}{5}{5} & \lomedup{32}{32}{32}\\
Red Wine &
\lomedup{25}{34}{55} & \lomedup{28}{28}{28} & \lomedup{11}{13}{19} & \lomedup{5}{\textbf{7}}{8}  & \lomedup{88}{88}{88}\\
Car Price & \lomedup{5}{5}{5} & \lomedup{28}{28}{28} & \lomedup{4}{\textbf{4}}{5} & \lomedup{5}{5}{5} & \lomedup{32}{32}{32}\\
Friedman1 & \lomedup{10}{12}{12} & \lomedup{28}{28}{28} & \lomedup{7}{7}{7} & \lomedup{6}{\textbf{6}}{6} & \lomedup{80}{80}{80}\\
Friedman2 &
\lomedup{5}{5}{5} & \lomedup{19}{19}{19} & \lomedup{4}{5}{5} & \lomedup{4}{\textbf{4}}{4} & \lomedup{32}{32}{32}\\
Friedman3 & \lomedup{7}{7}{7} & \lomedup{28}{28}{28} & \lomedup{11}{11}{11} & \lomedup{5}{\textbf{5}}{5} & \lomedup{32}{32}{32}\\
Magic & \lomedup{27}{27}{32} & \lomedup{\inft}{\inft}{\inft} & \lomedup{20}{27}{36} & \lomedup{11}{\textbf{12}}{12} & \lomedup{100}{100}{100}\\
Adult & \lomedup{24}{28}{28} & \lomedup{34}{\inft}{\inft} & \lomedup{12}{13}{17} & \lomedup{7}{\textbf{7}}{7} & \lomedup{\inft}{\inft}{\inft}\\
Voice & \lomedup{3}{\textbf{3}}{3} & \lomedup{51}{69}{82} & \lomedup{7}{9}{13} & \lomedup{4}{5}{5} & \lomedup{\inft}{\inft}{\inft}\\
Housing Price & \lomedup{3}{\textbf{5}}{5} & \lomedup{10}{10}{10} & \lomedup{7}{7}{7} & \lomedup{6}{6}{6} & \lomedup{64}{64}{64}\\
\botrule
\textbf{\# Wins} & 3 & 0 & 1 & 9 & 0\\
\midrule
& \multicolumn{5}{c}{50\% risk reduction} \\ \cmidrule{2-6}
Iris & \lomedup{8}{16}{23} & \lomedup{\inft}{\inft}{\inft} & \lomedup{20}{26}{33} & \lomedup{7}{\textbf{7}}{8} & \lomedup{40}{40}{40}\\
Liver & \lomedup{\inft}{\inft}{\inft} & \lomedup{\inft}{\inft}{\inft} & \lomedup{\inft}{\inft}{\inft} & \lomedup{\inft}{\inft}{\inft} & \lomedup{\inft}{\inft}{\inft}\\
Diabetes & 
\lomedup{\inft}{\inft}{\inft} & \lomedup{\inft}{\inft}{\inft} & \lomedup{\inft}{\inft}{\inft} & \lomedup{36}{\inft}{\inft} & \lomedup{\inft}{\inft}{\inft}\\
Breast Cancer & 
\lomedup{7}{7}{7} & \lomedup{36}{50}{66} & \lomedup{7}{7}{13} & \lomedup{5}{\textbf{5}}{5} & \lomedup{\inft}{\inft}{\inft}\\
Banknote & \lomedup{10}{10}{10} & \lomedup{48}{50}{54} & \lomedup{9}{9}{11} & \lomedup{5}{\textbf{5}}{5} & \lomedup{32}{32}{32}\\
Red Wine &
\lomedup{\inft}{\inft}{\inft} & \lomedup{\inft}{\inft}{\inft} & \lomedup{\inft}{\inft}{\inft} & \lomedup{\inft}{\inft}{\inft}  & \lomedup{\inft}{\inft}{\inft}\\
Car Price & \lomedup{5}{10}{12} & \lomedup{28}{28}{28} & \lomedup{4}{\textbf{4}}{5} & \lomedup{5}{5}{5} & \lomedup{32}{32}{32}\\
Friedman1 & \lomedup{25}{26}{28} & \lomedup{28}{28}{\inft} & \lomedup{7}{7}{7} & \lomedup{6}{\textbf{6}}{7} & \lomedup{80}{80}{80}\\
Friedman2 &
\lomedup{5}{5}{5} & \lomedup{19}{19}{19} & \lomedup{4}{5}{5} & \lomedup{4}{\textbf{4}}{4} & \lomedup{32}{32}{32}\\
Friedman3 & \lomedup{14}{14}{14} & \lomedup{28}{28}{28} & \lomedup{18}{28}{33} & \lomedup{11}{\textbf{12}}{12} & \lomedup{80}{80}{80}\\
Magic & \lomedup{\inft}{\inft}{\inft} & \lomedup{\inft}{\inft}{\inft} & \lomedup{\inft}{\inft}{\inft} & \lomedup{\inft}{\inft}{\inft} & \lomedup{\inft}{\inft}{\inft}\\
Adult & \lomedup{\inft}{\inft}{\inft} & \lomedup{\inft}{\inft}{\inft} & \lomedup{\inft}{\inft}{\inft} & \lomedup{\inft}{\inft}{\inft} & \lomedup{\inft}{\inft}{\inft}\\
Voice & \lomedup{3}{\textbf{3}}{3} & \lomedup{51}{69}{82} & \lomedup{7}{9}{13} & \lomedup{4}{5}{5} & \lomedup{\inft}{\inft}{\inft}\\
Housing Price & \lomedup{37}{38}{42} & \lomedup{\inft}{\inft}{\inft} & \lomedup{32}{35}{45} & \lomedup{13}{\textbf{14}}{14} & \lomedup{\inft}{\inft}{\inft}\\
\botrule
\textbf{\# Wins} & 1 & 0 & 1 & 7 & 0\\
\midrule
& \multicolumn{5}{c}{75\% risk reduction} \\ \cmidrule{2-6}
Iris & 
\lomedup{35}{44}{\inft} & \lomedup{\inft}{\inft}{\inft} & \lomedup{69}{76}{102} & \lomedup{8}{\textbf{8}}{15} & \lomedup{80}{80}{\inft}\\
Liver & \lomedup{\inft}{\inft}{\inft} & \lomedup{\inft}{\inft}{\inft} & \lomedup{\inft}{\inft}{\inft} & \lomedup{\inft}{\inft}{\inft} & \lomedup{\inft}{\inft}{\inft}\\
Diabetes & 
\lomedup{\inft}{\inft}{\inft} & \lomedup{\inft}{\inft}{\inft} & \lomedup{\inft}{\inft}{\inft} & \lomedup{\inft}{\inft}{\inft} & \lomedup{\inft}{\inft}{\inft}\\
Breast Cancer & 
\lomedup{50}{69}{102} & \lomedup{63}{66}{\inft} & \lomedup{13}{20}{32} & \lomedup{6}{\textbf{7}}{8} & \lomedup{\inft}{\inft}{\inft}\\
Banknote & \lomedup{31}{33}{38} & \lomedup{\inft}{\inft}{\inft} & \lomedup{9}{9}{11} & \lomedup{5}{\textbf{5}}{5} & \lomedup{40}{40}{40}\\
Red Wine &
\lomedup{\inft}{\inft}{\inft} & \lomedup{\inft}{\inft}{\inft} & \lomedup{\inft}{\inft}{\inft} & \lomedup{\inft}{\inft}{\inft}  & \lomedup{\inft}{\inft}{\inft}\\
Car Price & \lomedup{17}{19}{24} & \lomedup{28}{28}{28} & \lomedup{64}{104}{\inft} & \lomedup{13}{\textbf{13}}{14} & \lomedup{40}{40}{40}\\
Friedman1 & \lomedup{74}{80}{100} & \lomedup{\inft}{\inft}{\inft} & \lomedup{\inft}{\inft}{\inft} & \lomedup{50}{\textbf{54}}{57} & \lomedup{100}{100}{100}\\
Friedman2 &
\lomedup{13}{\textbf{13}}{13} & \lomedup{19}{19}{19} & \lomedup{23}{27}{33} & \lomedup{12}{\textbf{13}}{13} & \lomedup{40}{40}{40}\\
Friedman3 & \lomedup{\inft}{\inft}{\inft} & \lomedup{\inft}{\inft}{\inft} & \lomedup{\inft}{\inft}{\inft} & \lomedup{54}{64}{74} & \lomedup{\inft}{\inft}{\inft}\\
Magic & \lomedup{\inft}{\inft}{\inft} & \lomedup{\inft}{\inft}{\inft} & \lomedup{\inft}{\inft}{\inft} & \lomedup{\inft}{\inft}{\inft} & \lomedup{\inft}{\inft}{\inft}\\
Adult & \lomedup{\inft}{\inft}{\inft} & \lomedup{\inft}{\inft}{\inft} & \lomedup{\inft}{\inft}{\inft} & \lomedup{\inft}{\inft}{\inft} & \lomedup{\inft}{\inft}{\inft}\\
Voice & \lomedup{10}{15}{17} & \lomedup{51}{69}{82} & \lomedup{7}{9}{13} & \lomedup{4}{\textbf{5}}{5} & \lomedup{\inft}{\inft}{\inft}\\
Housing Price & \lomedup{\inft}{\inft}{\inft} & \lomedup{\inft}{\inft}{\inft} & \lomedup{\inft}{\inft}{\inft} & \lomedup{\inft}{\inft}{\inft} & \lomedup{\inft}{\inft}{\inft}\\
\botrule
\textbf{\# Wins} & 1 & 0 & 0 & 7 & 0\\
\end{tabular*}
\end{table}
\begin{table}[t]
\caption{Normalized risk comparison of methods across benchmark datasets, averaged over complexities between 1 to 100 based on \textbf{oracle-selected} hyper-parameters. For cases where a method does not achieve a valid solution within this complexity budget, (–) is reported.}\label{table:risk_oracle}
\centering
\begin{tabular*}{\textwidth}{@{\extracolsep\fill}lccccc}
\toprule%
& \multicolumn{5}{c}{Train Risks} \\ \cmidrule{2-6}
Dataset & TGB	& NRE & RFObTr & LLTB & EBM \\
\midrule
Iris	& 0.25 \extratiny{$\pm$ 0.09} & 0.45 \extratiny{$\pm$ 0.05} & 0.19 \extratiny{$\pm$ 0.10} & \textbf{0.08} \extratiny{$\pm$ 0.02} & 0.51 \extratiny{$\pm$ 0.06}\\
Liver	& 0.77 \extratiny{$\pm$ 0.04} & 0.83 \extratiny{$\pm$ 0.00} & \textbf{0.70} \extratiny{$\pm$ 0.03} & 0.79 \extratiny{$\pm$ 0.03} & 0.84 \extratiny{$\pm$ 0.00}\\
Diabetes	& \textbf{0.45} \extratiny{$\pm$ 0.05} & 0.46 \extratiny{$\pm$ 0.02} & 0.46 \extratiny{$\pm$ 0.02} & 0.48 \extratiny{$\pm$ 0.03} & 0.53 \extratiny{$\pm$ 0.00} \\
Breast Cancer	& 0.18 \extratiny{$\pm$ 0.04} & 0.18 \extratiny{$\pm$ 0.03} & \textbf{0.06} \extratiny{$\pm$ 0.01} & 0.09 \extratiny{$\pm$ 0.02} & -\\
Banknote	& 0.23 \extratiny{$\pm$ 0.10} & 0.31 \extratiny{$\pm$ 0.02} & \textbf{0.01} \extratiny{$\pm$ 0.00} & \textbf{0.01} \extratiny{$\pm$ 0.01} & 0.35 \extratiny{$\pm$ 0.06} \\
Red Wine	& \textbf{0.63} \extratiny{$\pm$ 0.04} & 0.65 \extratiny{$\pm$ 0.01} & \textbf{0.63} \extratiny{$\pm$ 0.02} & \textbf{0.63} \extratiny{$\pm$ 0.02} & 0.72 \extratiny{$\pm$ 0.00}\\
Car Price	& 0.18 \extratiny{$\pm$ 0.06} & \textbf{0.09} \extratiny{$\pm$ 0.00} & 0.28 \extratiny{$\pm$ 0.02} & 0.13 \extratiny{$\pm$ 0.04} & 0.28 \extratiny{$\pm$ 0.02} \\
Friedman1	& 0.37 \extratiny{$\pm$ 0.10} & 0.42 \extratiny{$\pm$ 0.00} & 0.42 \extratiny{$\pm$ 0.02} & \textbf{0.26} \extratiny{$\pm$ 0.06} & 0.34 \extratiny{$\pm$ 0.00}\\
Friedman2	& 0.12 \extratiny{$\pm$ 0.06} & \textbf{0.02} \extratiny{$\pm$ 0.00} & 0.22 \extratiny{$\pm$ 0.02} & 0.08 \extratiny{$\pm$ 0.04} & 0.26 \extratiny{$\pm$ 0.01}\\
Friedman3	& 0.36 \extratiny{$\pm$ 0.04} & 0.48 \extratiny{$\pm$ 0.05} & 0.36 \extratiny{$\pm$ 0.04} & \textbf{0.29} \extratiny{$\pm$ 0.05} & 0.57 \extratiny{$\pm$ 0.04} \\
Magic	& 0.69 \extratiny{$\pm$ 0.05} & \textbf{0.61} \extratiny{$\pm$ 0.00} & 0.70 \extratiny{$\pm$ 0.03} & \textbf{0.61} \extratiny{$\pm$ 0.04} & 0.73 \extratiny{$\pm$ 0.00}\\
Adult	& 0.70 \extratiny{$\pm$ 0.03} & 0.77 \extratiny{$\pm$ 0.06} & 0.66 \extratiny{$\pm$ 0.04} & \textbf{0.61} \extratiny{$\pm$ 0.03} & -\\
Voice	& 0.17 \extratiny{$\pm$ 0.05} & 0.16 \extratiny{$\pm$ 0.00} & \textbf{0.12} \extratiny{$\pm$ 0.01} & \textbf{0.12} \extratiny{$\pm$ 0.02} & -\\
Housing Price	& 0.47 \extratiny{$\pm$ 0.05} & 0.47 \extratiny{$\pm$ 0.02} & 0.48 \extratiny{$\pm$ 0.02} & \textbf{0.36} \extratiny{$\pm$ 0.04} & 0.59 \extratiny{$\pm$ 0.00}\\
\midrule
& \multicolumn{5}{c}{Test Risks} \\ \cmidrule{2-6}
Iris	& 0.39 \extratiny{$\pm$ 0.08} & 0.53 \extratiny{$\pm$ 0.02} & 0.37 \extratiny{$\pm$ 0.09} & \textbf{0.24} \extratiny{$\pm$ 0.01} & 0.66 \extratiny{$\pm$ 0.08}\\
Liver	& 0.93 \extratiny{$\pm$ 0.01} & 0.96 \extratiny{$\pm$ 0.00} & 0.90 \extratiny{$\pm$ 0.01} & \textbf{0.87} \extratiny{$\pm$ 0.01} & 0.89 \extratiny{$\pm$ 0.00} \\
Diabetes	& 0.70 \extratiny{$\pm$ 0.02} & 0.59 \extratiny{$\pm$ 0.01} & 0.61 \extratiny{$\pm$ 0.02} & \textbf{0.55} \extratiny{$\pm$ 0.01} & 0.57 \extratiny{$\pm$ 0.00}\\
Breast Cancer	& 0.27 \extratiny{$\pm$ 0.03} & 0.21 \extratiny{$\pm$ 0.03} & 0.19 \extratiny{$\pm$ 0.02} & \textbf{0.18} \extratiny{$\pm$ 0.01} & -\\
Banknote	& 0.25 \extratiny{$\pm$ 0.09} & 0.31 \extratiny{$\pm$ 0.03} & 0.02 \extratiny{$\pm$ 0.01} & \textbf{0.01} \extratiny{$\pm$ 0.01} & 0.36 \extratiny{$\pm$ 0.06}\\
Red Wine	& 0.74 \extratiny{$\pm$ 0.02} & 0.69 \extratiny{$\pm$ 0.00} & 0.69 \extratiny{$\pm$ 0.01} & \textbf{0.66} \extratiny{$\pm$ 0.01} & 0.73 \extratiny{$\pm$ 0.00} \\
Car Price	& 0.19 \extratiny{$\pm$ 0.06} & \textbf{0.09} \extratiny{$\pm$ 0.00} & 0.29 \extratiny{$\pm$ 0.02} & 0.14 \extratiny{$\pm$ 0.04} & 0.28 \extratiny{$\pm$ 0.02}\\
Friedman1	& 0.39 \extratiny{$\pm$ 0.09} & 0.44 \extratiny{$\pm$ 0.00} & 0.44 \extratiny{$\pm$ 0.02} & \textbf{0.27} \extratiny{$\pm$ 0.05} &0.34 \extratiny{$\pm$ 0.00} \\
Friedman2	& 0.12 \extratiny{$\pm$ 0.06} & \textbf{0.02} \extratiny{$\pm$ 0.00} & 0.23 \extratiny{$\pm$ 0.02} & 0.08 \extratiny{$\pm$ 0.04} & 0.25 \extratiny{$\pm$ 0.01}\\
Friedman3	& 0.37 \extratiny{$\pm$ 0.04} & 0.50 \extratiny{$\pm$ 0.06} & 0.37 \extratiny{$\pm$ 0.04} & \textbf{0.30} \extratiny{$\pm$ 0.05} & 0.58 \extratiny{$\pm$ 0.04}\\
Magic	& 0.71 \extratiny{$\pm$ 0.04} & \textbf{0.63} \extratiny{$\pm$ 0.00} & 0.72 \extratiny{$\pm$ 0.02} & \textbf{0.63} \extratiny{$\pm$ 0.03} & 0.74 \extratiny{$\pm$ 0.00} \\
Adult	& 0.72 \extratiny{$\pm$ 0.02} & 0.77 \extratiny{$\pm$ 0.06} & 0.68 \extratiny{$\pm$ 0.04} & \textbf{0.63} \extratiny{$\pm$ 0.02} & - \\
Voice	& 0.19 \extratiny{$\pm$ 0.04} & 0.16 \extratiny{$\pm$ 0.00} & 0.15 \extratiny{$\pm$ 0.01} & \textbf{0.13} \extratiny{$\pm$ 0.01} & -\\
Housing Price	& 0.49 \extratiny{$\pm$ 0.04} & 0.47 \extratiny{$\pm$ 0.02} & 0.49 \extratiny{$\pm$ 0.02} & \textbf{0.38} \extratiny{$\pm$ 0.04} & 0.59 \extratiny{$\pm$ 0.00}\\
\botrule
\end{tabular*}
\end{table}
\subsection{Bisection Search Computational Complexity}\label{sec_bisection_search}
\begin{figure}[t]
    \centering
    \includegraphics[width=1\linewidth]{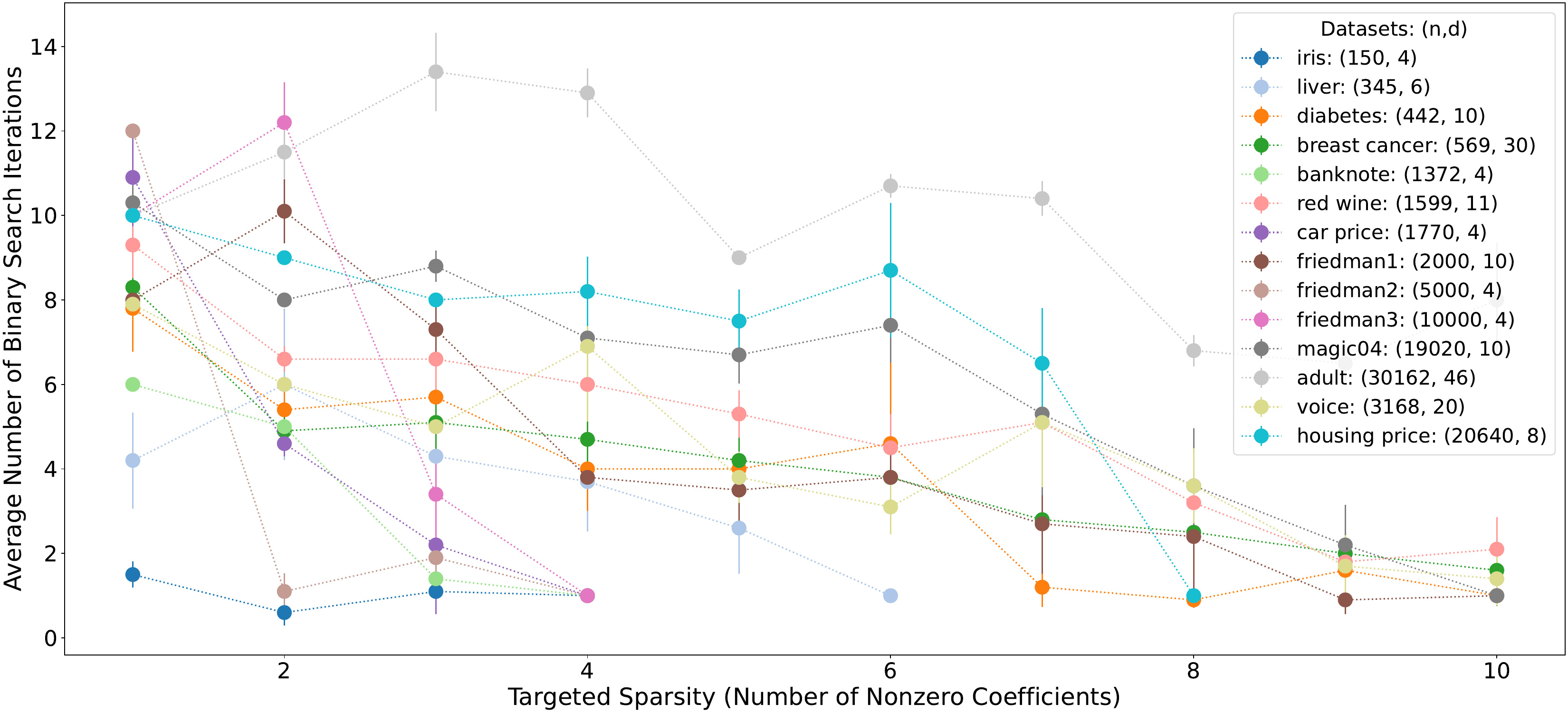}
    \caption{Empirical computational complexity of the binary search procedure across 14 benchmark datasets with original sample sizes. The x-axis denotes the target sparsity level (up to 10), and the y-axis shows the average number of binary search iterations required to identify the $\lambda_{l1}$ value yielding the desired sparsity. Error bars indicate 95\% confidence intervals over 10 repetitions.}
    \label{fig:binary_search_comp_comp}
\end{figure}

The proposed method uses bisection search to find $\lambda_s$ from the interval $I_s = [a_s, b_s]$ of regularization values that yield an $s$-sparse solution to the L1-regularized weighted logistic regression problem given in \Cref{eq:l1logreg}.
\Cref{fig:binary_search_comp_comp} illustrates the number of bisection search iterations $B$ required across benchmark datasets to achieve the different target sparsity levels $s$. 
We see an overall decreasing trend of the number of required iterations in terms of $s$, no systematic dependency on $n$, and only a very mild increasing dependency on $d$---suggesting that the factor $B$ in the complexity \Cref{eq:comp_complexity} is overall well behaved and can largely be treated as a constant.
Note that, in this analysis, the original sample size of each dataset is used. For datasets with categorical variables, such as \texttt{Adult}, the reported number of features corresponds to the dimensionality after one-hot encoding.
\section{Distribution-free Confidence Interval for Median}\label{sec_median_ci}
The median confidence intervals follow a well-developed nonparametric theory~\citep[see, e.g, Ch. 5][]{hahn2011statistical} based on order statistics and binomial tail probabilities, which is particularly advantageous in our setting because it yields distribution-free coverage guarantees and remains robust in the presence of censored observations (e.g., runs complexities recorded as $\infty$), where mean- and variance-based summaries can be unstable.

Let $X_1,\dots,X_n$ be draws from a distribution $F$, and let $X_{(1)} \le \cdots \le X_{(n)}$ denote the ordered values (order statistics). Let $F_{1/2}$ denote the population median of $F$. Under continuity of $F$, we have $\Pr(X_i \le F_{1/2}) = 1/2$, so the number of observations below (or equivalently, not exceeding) the true median follows a $\mathrm{Binomial}(n,1/2)$ law. Consequently, the coverage probability of the random interval $[X_{(l)},X_{(u)}]$ depends only on $(n,l,u)$ and is
\begin{equation*}
\Pr\!\big(X_{(l)} \le F_{1/2} \le X_{(u)}\big)
= 2^{-n}\sum_{j=l}^{u}\binom{n}{j}.
\end{equation*}
By choosing integers $l$ and $u$ such that $2^{-n}\sum_{j=l}^{u}\binom{n}{j} \ge 1-\alpha$, the interval $[X_{(l)},X_{(u)}]$ is a distribution-free confidence interval for the median with confidence level at least $1-\alpha$. For example, with $n=15$, selecting $l=5$ and $u=11$ yields coverage $\approx 0.923$, i.e., a $\sim 92\%$ confidence interval $[X_{(5)},X_{(11)}]$.
\end{appendices}
\end{document}